\documentclass[times,twocolumn,final]{elsarticle}
\usepackage{medima}
\usepackage{framed,multirow}
\usepackage{tabularray}
\usepackage{amssymb}
\usepackage{latexsym}
\usepackage{url}
\usepackage{xcolor}
\usepackage{amsmath}
\usepackage{hyperref}
\usepackage{makecell}
\definecolor{newcolor}{rgb}{.8,.349,.1}
\begin{document}
% \verso{Yunyi Liu \textit{et~al.}}
\begin{frontmatter}
\title{A Systematic Evaluation of GPT-4V's Multimodal Capability for Medical Image Analysis}
\author[1]{Yunyi \snm{Liu}\corref{tnote1}}
\author[1]{Yingshu \snm{Li}\corref{tnote1}}
\author[1,4]{Zhanyu \snm{Wang}\corref{tnote1}}
\author[5]{Xinyu \snm{Liang}}
\author[2]{Lei \snm{Wang}}
\author[3]{Lingqiao \snm{Liu}}
\author[4]{Leyang \snm{Cui}}
\author[4]{Zhaopeng \snm{Tu}}
\author[4]{Longyue \snm{Wang}\corref{cor1}}
\ead{vinnylywang@tencent.com}
\cortext[tnote1]{Equal Contribution}
\cortext[cor1]{Corresponding author}
\author[1]{Luping \snm{Zhou}\corref{cor1}}\ead{luping.zhou@sydney.edu.au}
\address[1]{Electrical and Computer Engineering, The University of Sydney, NSW, 2006, Australia}
\address[2]{School of Computing and Information Technology, The University of Wollongong, NSW, 2522, Australia}
\address[3]{School of Computer and Mathematical Sciences, The University of Adelaide, Adelaide, 5005, Australia}
\address[4]{Tencent AI Lab, Tencent, Shenzhen, 518000, China}
\address[5]{First Clinical Medical College, Guangzhou University of Chinese Medicine, Guangzhou, 510405, China}
\begin{abstract}
This work conducts an evaluation of GPT-4V's multimodal capability for medical image analysis, with a focus on three representative tasks of radiology report generation, medical visual question answering, and medical visual grounding. For the evaluation, a set of prompts is designed for each task to induce the corresponding capability of GPT-4V to produce sufficiently good outputs. Three evaluation ways including quantitative analysis, human evaluation, and case study are employed to achieve an in-depth and extensive evaluation. Our evaluation shows that GPT-4V excels in understanding medical images and is able to generate high-quality radiology reports and effectively answer questions about medical images. Meanwhile, it is found that its performance for medical visual grounding needs to be substantially improved. In addition, we observe the discrepancy between the evaluation outcome from quantitative analysis and that from human evaluation. This discrepancy suggests the limitations of conventional metrics in assessing the performance of large language models like GPT-4V and the necessity of developing new metrics for automatic quantitative analysis.
\end{abstract}
\end{frontmatter}
%% main text
\section{Introduction}
\label{sec1}
%\noindent\textbf{Large Language Models (LLMs)} LLMs have consistently demonstrated remarkable ability across various domains and tasks~\citep{touvron2023llama, OpenAI2023GPT4TR, anil2023palm}. The ongoing pursuit of enhancing LLMs' capacity for visual comprehension has spurred the emergence of a new research area: Large Multi-modal Models (LMMs)~\citep{ye2023mplug, li2023blip, awadalla2023openflamingo}. The basic approach has been to either fine-tune the visual encoder to align with a fixed pre-trained LLM or to use a vision-language model to convert visual input into textual descriptions that can be understood by the LLM. These applications are all based solely on the use of the LLM and do not really explore the visual capabilities of the LLM. GPT-4V, a cutting-edge Large Multi-modal Model (LMM) incorporating visual understanding capabilities, is constructed as an evolution of state-of-the-art Large Multi-modal Models (LMMs). This model is trained on an extensive corpus of multi-modal data. \cite{yang2023dawn} conducted a comprehensive case study to assess GPT-4V's performance in general-purpose scenarios, revealing its robust visual comprehension ability~\citep{yang2023dawn}. Meanwhile, LMMs have been widely used in the medical field~\citep{wang2023r2gengpt, singhal2023towards}. The introduction of visual capabilities into GPT-4V opens up opportunities for an in-depth examination of its potential in the domain of medical multi-modality. 

\noindent\textbf{Large language models (LLMs)} have recently demonstrated remarkable ability across various domains and tasks~\citep{touvron2023llama, OpenAI2023GPT4TR, anil2023palm}. The ongoing pursuit of enhancing LLMs' capability for visual comprehension has further spurred the emergence of large multimodal models (LMMs)~\citep{ye2023mplug, li2023blip, awadalla2023openflamingo}. 
%The basic approach has been to either fine-tune the visual encoder to align with a fixed pre-trained LLM or to use a vision-language model to convert visual input into textual descriptions that can be understood by the LLM. These applications are all based solely on the use of the LLM and do not really explore the visual capabilities of the LLM. 
Among them, GPT-4V incorporates visual understanding capability by learning from an extensive corpus of multimodal data and demonstrates great potential on multimodal data comprehension. It is a state-of-the-art model known for its proficiency in image analysis and text generation. To have a precise understanding of GPT-4V's potential, 
%systematic evaluation of its performance on various tasks has attracted research attention. 
research has been newly conducted to investigate its capability for generic images~\citep{wu2023early}. For example, \cite{yang2023dawn} conducted a case study to assess GPT-4V's performance in general-purpose scenarios, revealing its robust visual comprehension ability. 

LLMs could have enormous applications to medicine and healthcare~\citep{wang2023r2gengpt, singhal2023towards}. The availability of vision module in GPT-4V opens up an opportunity for an in-depth examination of its potential in this regard. Some papers have started examining its performance on medical images~\citep{wu2023can}. However, they are primarily based on case studies. 

In this work, we move beyond case studies and delve deeper into the capability of GPT-4V for multimodal tasks in medical image analysis. Our evaluations focus on three representative tasks in this regard, including radiology report generation, medical visual question answering, and medical visual grounding. This enables us to assess GPT-4V's ability in understanding and interacting between visual and textual modalities from various perspectives. Our evaluation takes a multifaceted approach consisting of quantitative analysis, human evaluations, and case studies. Quantitative analysis calculates conventional performance-related metrics and compares GPT-4V with other relevant models by following the protocols in the literature. Human evaluation is used to obtain a more precise assessment of the quality of text output of GPT-4V in the tasks of radiology report generation and medical visual question answering. This is particularly meaningful when considering that conventional metrics are not able to provide a direct and whole assessment. Case study is utilised to facilitate an intuitive understanding and assessment of the output. Altogether, the three evaluation methods offer an extensive view of the performance of GPT-4V.

Various prompt settings are explored in the evaluation of GPT-4V. Zero-shot prompts and few-shot prompts are tested for the task of radiology report generation based on whether a set of example reference reports is provided. For medical visual question answering, several examples are provided as prompts to guide GPT-4V to respond in a way that better aligns with the form of ground truth. For medical visual grounding, the prompt is mainly used to inform GPT-4V of the specific requirement of the output, particularly on creating the bounding boxes for visual grounding. Evaluation with the above prompt settings allows the capability of GPT-4V to be properly activated and its performance concerning prompt to be assessed.  

\noindent The results of our evaluations are summarised as follows.
\begin{itemize}
\item[$\bullet$] 
GPT-4V demonstrates promising performance in generating radiology reports for medical images. Assessed by either language fluency related metrics or clinical efficacy related ones, its performance is competitive with that of the models specially developed for this task. In addition, compared with the conventional metrics used in quantitative analysis, the human evaluation suggests an even higher level of accuracy, relevancy, and richness of the radiology reports generated by GPT-4V. 
%This discrepancy underscores the limitations of conventional metrics in assessing the outputs of large language models (LLMs) like GPT-4V.

\item[$\bullet$] 
GPT-4V shows its ability in generating detailed answers for the task of medical visual question answering. Meanwhile, when compared with the relevant state-of-the-art methods, it only achieves modest performance when assessed by the conventional metrics. This is attributed to the low correlation between questions and the generated answers, the great diversity of the answers generated by GPT-4V, and the fixed format of the ground-truth answers in the benchmark dataset. Meanwhile, human evaluation suggests that GPT-4V's answers are more accurate than what is indicated by the conventional metrics.

\item[$\bullet$] 
The result of the medical visual grounding task highlights a significant room for improvement. GPT-4V struggles with accurately locating and identifying specific elements within medical images. This is consistent with the weakness revealed by existing evaluation on the tasks related to generic images. This finding points out a potential direction to further enhance GPT-4V’s performance.

\item[$\bullet$] 
In addition to utilising conventional metrics, our evaluation also involves professional medical practitioners conducting human evaluation, adding a layer of expert review to our assessment. Their discrepancy reveals the limitations of conventional performance-related metrics in accurately assessing the performance of GPT-4V, highlighting the urgency of developing more proper ways of evaluation for LLMs.
\end{itemize}

\noindent The remainder of the paper is organized as follows. Section 2 is the related work for GPT-4V. Section 3 introduces our Evaluation, which contains evaluation tasks, evaluation method, evaluation datasets and evaluation results. Section 4 describes our summary and discussion. We conclude the paper and discuss the limitation in Section 5.

% Please use \verb+elsarticle.cls+ for typesetting your paper.
% Additionally load the package \verb+medima.sty+ in the preamble using
% the following command: 
% \begin{verbatim} 
%   \usepackage{medima}
% \end{verbatim}

% Following commands are defined for this journal which are not in
% \verb+elsarticle.cls+. 
% \begin{verbatim}
%   \received{}
%   \finalform{}
%   \accepted{}
%   \availableonline{}
%   \communicated{}
% \end{verbatim}

% Any instructions relavant to the \verb+elsarticle.cls+ are applicable
% here as well. See the online instruction available on:
% \makeatletter
% \if@twocolumn
% \begin{verbatim}
%  http://support.stmdocs.com/wiki/
%  index.php?title=Elsarticle.cls
% \end{verbatim}
% \else
% \begin{verbatim}
%  http://support.stmdocs.com/wiki/index.php?title=Elsarticle.cls
% \end{verbatim}
% \fi

% \subsection{Entering text}
% \textcolor{newcolor}{\bf There is no page limit.}
\section{GPT-4V~\citep{OpenAI2023GPT4TR}}

 GPT-4V, or GPT-4 with Vision, stands as an advanced multimodal system developed by OpenAI, integrating image inputs into LLMs. This groundbreaking fusion represents a notable frontier in AI research, enabling innovative interfaces and the solution of new tasks for unique user experiences. The system is built upon the GPT-4 model and has been trained on an extensive dataset comprising text and image data from the internet and licensed sources. GPT-4V showcases what can and cannot be done with text and image processing, highlighting new strengths that come from combining these two types of data. This includes advanced intelligence and the ability to reason across a broad range of topics.
 %GPT-4V presents both the limitations and the capabilities of text and vision modalities and introduces new capabilities emerging from their intersection, including intelligence and reasoning at a large scale. 

Several recent publications have explored the multimodal capacities of GPT-4V. Early evaluations, as presented in~\citep{wu2023early}, offered insights into GPT-4V’s abilities and limitations. \citep {li2023comprehensive} focused on Visual Question Answering (VQA) tasks, while~\citep{shi2023exploring} explored GPT-4V's Optical Character Recognition (OCR) capabilities. Some research works, like~\citep{wu2023can}, specifically investigated medical image tasks, albeit with a more case-study-oriented approach lacking in-depth exploration. In contrast, our paper aims to provide a deeper study of GPT-4V’s multimodal capabilities in medical imaging contexts.

\section{Evaluation}

The upcoming paragraphs provide a detailed introduction to four key components of our evaluation process.

\noindent\textbf{Evaluation Tasks} describes the specific tasks that we set up for the evaluation.

\noindent\textbf{Evaluation Method} outlines how we design the prompts and describes the methodology used to assess the models. This includes the criteria for evaluation, the metrics such as accuracy, precision, recall, F1 score, BLEU, CIDEr, ROUGE, METEOR, accuracy, and mIoU, and any specific protocols or procedures that we followed during the evaluation process.

\noindent\textbf{Evaluation Datasets} provides details about the datasets involved in the evaluation. Important aspects like the source, the sizes, and the characteristics of the datasets, as well as their relevance to the evaluation tasks, are discussed. 

\noindent\textbf{Evaluation Results} presents the results of the evaluation. It includes a summary of the findings, comparisons with benchmarks or previous models (if applicable), and insights or interpretations of the results. It is partitioned into three parts: the quantitative, human evaluation, and case study results.

\subsection{Evaluation Tasks}
\begin{figure}[h]
\begin{center}
%\framebox[4.0in]{$\;$}
\includegraphics[width=0.45\textwidth]{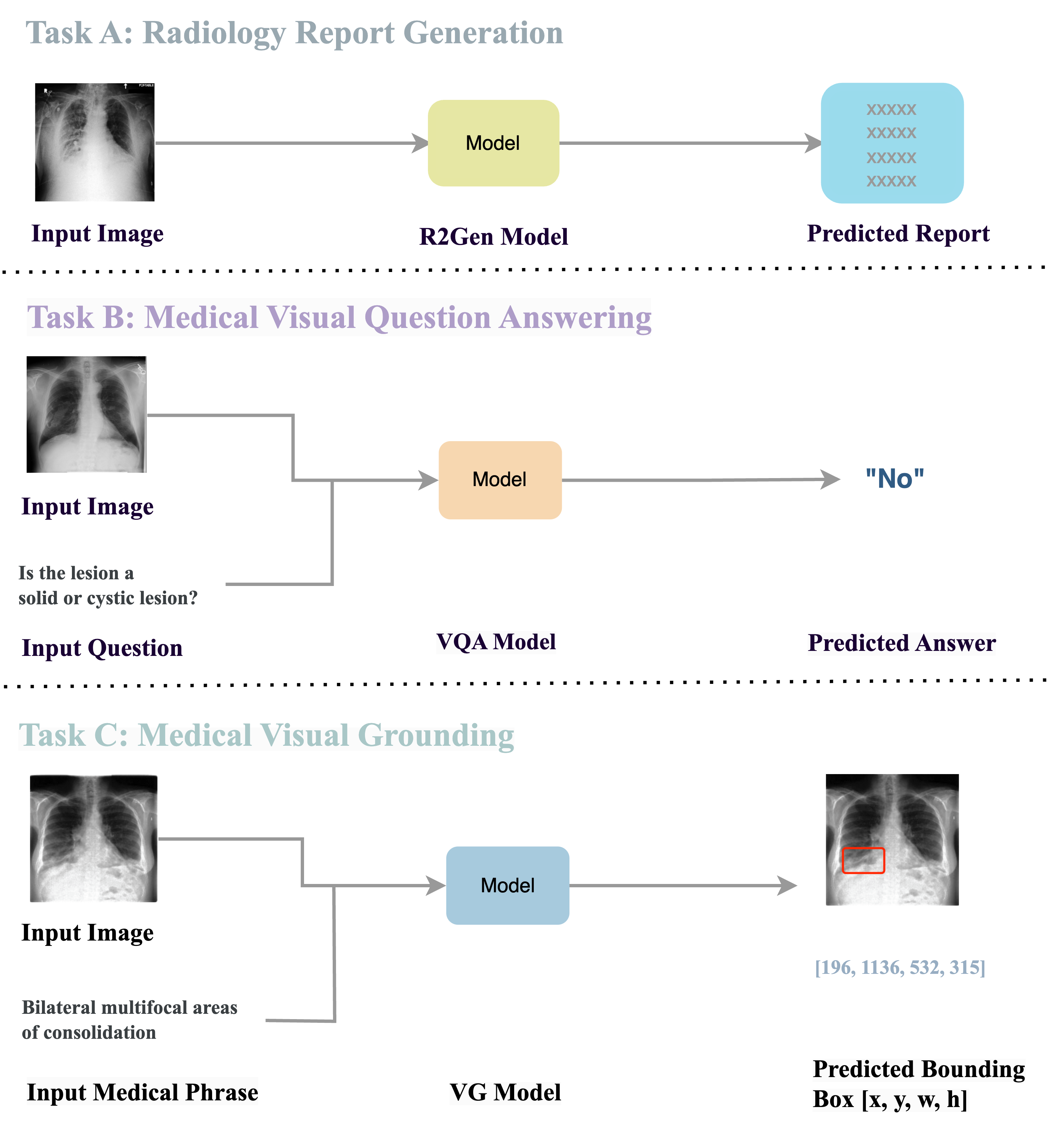}
\end{center}
\vspace{-5mm}
\caption{Three multimodal medical imaging tasks we employ to evaluate GPT-4V's performance.}
\label{fig: evaluationTasksIntro}
\end{figure}

\noindent In the following, we introduce the three evaluation tasks shown in Figure \ref{fig: evaluationTasksIntro}, including radiology report generation (R2Gen), medical Visual Question Answering (VQA), and medical visual grounding (VG). 

\subsubsection{Radiology Report Generation} 
\noindent Radiology report generation (R2Gen) is a very important application of medical images, akin to the task of image captioning~\citep{vinyals2015showandtell, xu2016showattandtell, pan2020xlinear}. R2Gen poses unique challenges, given the inherent complexity of medical reports, their length, and the difficulty in discerning fine-grained abnormalities from medical images, particularly in datasets biased towards normal samples (referred to as data bias problem). Current research can be grouped into two primary research directions. The first direction concentrates on enhancing the model's architecture to facilitate improved extraction of visual features and the generation of high-quality medical reports. For example, \cite{li2018hybrid} used a hierarchical architecture to generate reports with normality and abnormality respectively. Building on the Transformer's success~\citep{vaswani2017attentionisallyouneed}, \cite{chen2020generating} introduced a Transformer-based model, enhancing it with relational memory and memory-driven conditional layer normalization to enhance image feature representation and capture crucial report patterns~\citep{chen2020generating}. The second research direction addresses the data bias problem by incorporating external knowledge information. For example, some works constructed predefined medical knowledge graphs to augment the model's ability to capture valuable clinical information~\citep{2020When, liu2021exploring, li2023dynamic, huang2023kiut}. 
% To further enrich this supplementary knowledge, \cite{li2023dynamic} developed a dynamic approach that enables real-time updates to the knowledge graph~\citep{li2023dynamic}. 
Furthermore, very recently, there has been a surge in radiology report generation methods leveraging Large Language Models (LLMs). These approaches harness the capabilities of LLM to generate long-text content and utilise abundant knowledge sources to enhance the quality of radiology reports. For example, \cite{wang2023r2gengpt} employed LLaMA2~\citep{touvron2023llama} to elevate the quality of the generated reports, ensuring effective image-text alignment through a visual mapper. 

\subsubsection{Visual Question Answering} 
\noindent The Visual Question Answering (VQA) task~\citep{jiang2020defense, wu2019differential} involves processing the input image-question pairs to generate appropriate answers. Currently, there are two predominant approaches for implementing VQA tasks: classification-based  ~\citep{nguyen2019overcoming,finn2017model,eslami2021does}, and generation-based ~\citep{ambati2018sequence,khare2021mmbert}. By nature, VQA should be based on generation. The ongoing shift from a classification-centric paradigm to a generation-oriented approach represents a prevailing trend in the VQA field. Based on the characteristics of the VQA task, a proficient text generation model is essential. Consequently, the current surge in LLMs presents a significant opportunity for substantial improvements in the VQA task. Numerous endeavors incorporating LLMs into VQA tasks are already underway, whether for generating VQA datasets~\citep{pellegrini2023rad} or utilising LLMs to enhance the performance of VQA systems~\citep{li2023llava}. The evident improvements that LLMs bring to VQA lead us to believe that they are well-suited for this task. Consequently, evaluating the VQA task is a crucial aspect of GPT-4V's evaluation.

\subsubsection{Visual Grounding} 
\noindent In the visual grounding (VG)~\citep{kamath2021mdetr} task, the input typically comprises a medical image accompanied by a descriptive statement about the image, often pertaining a specific medical sign or symptom. The task's output is the coordinates of a bounding box that visually marks the area described in the statement, such as encapsulating a particular medical sign. Most visual grounding research focuses on general images, with only a few studies targeting medical images, likely due to the scarcity of corresponding medical datasets. However, the recently introduced MS-CXR dataset has opened up new possibilities in medical visual grounding, leading to emerging publications~\citep{huang2023enhancing, sun2023you, sun2023scoping} based on this dataset. Despite growing recognition, there remains untapped potential, presenting a significant opportunity for future research in medical visual grounding. Unlike the previous two tasks, the output for visual grounding is not a conventional text paragraph but a set of coordinates. Recent studies have successfully integrated LLMs~\citep{peng2023kosmos, zhao2023bubogpt} to directly produce these coordinates as outputs, showing promising results. Recognizing this potential, we hypothesize that GPT-4V should also possess visual grounding capabilities. Consequently, we include this task in our evaluations to assess GPT-4V's performance in this specific area.

\subsection{Evaluation Method/Process }
\subsubsection {Radiology Report Generation}
\noindent To better activate the capabilities of GPT-4V, we explore various prompt design strategies, including the \textbf{zero-shot} and \textbf{few-shot} approaches. \\ \\
\noindent \textbf{Zero-shot Prompt:} In zero-shot scenario, we provide a prompt without reference reports, allowing GPT-4V to autonomously generate reports without external guidance. GPT-4V is tasked with generating both the ``impression" and ``findings" sections and compared with the ground truth report. 

\noindent \textbf{Few-shot Prompts:} In-context few-shot learning represents a crucial methodology for enhancing the capabilities of LLMs~\citep{tsimpoukelli2021multimodal, wei2022emergent, dai2022can}. It enables the model to acquire the expected output format by providing a set of examples. In contrast to fine-tuning, this method empowers the model to generate desired results without any parameter updating at inference time. In our investigation, we experiment with a few prompt strategies designed for GPT-4V. Specifically, we explore diverse compositions within the few-shot prompts:
\begin{itemize}
    \item Exclusively using normal examples \textbf{(Few-shot normal-example prompt)},
    \item Exclusively using abnormal examples \textbf{(Few-shot abnormal-example prompt)}, and
    \item Combining one normal and one abnormal example \textbf{(Few-shot mixed-example prompt)}.
\end{itemize}

% These scenarios allowed us to comprehensively assess GPT-4V's performance in radiology report generation, considering both its autonomous report generation ability and its ability to generate reports aligned with the Ground Truth format. 

\begin{figure}[h]
\centering
\begin{center}
%\framebox[4.0in]{$\;$}
\includegraphics[width=0.45\textwidth]{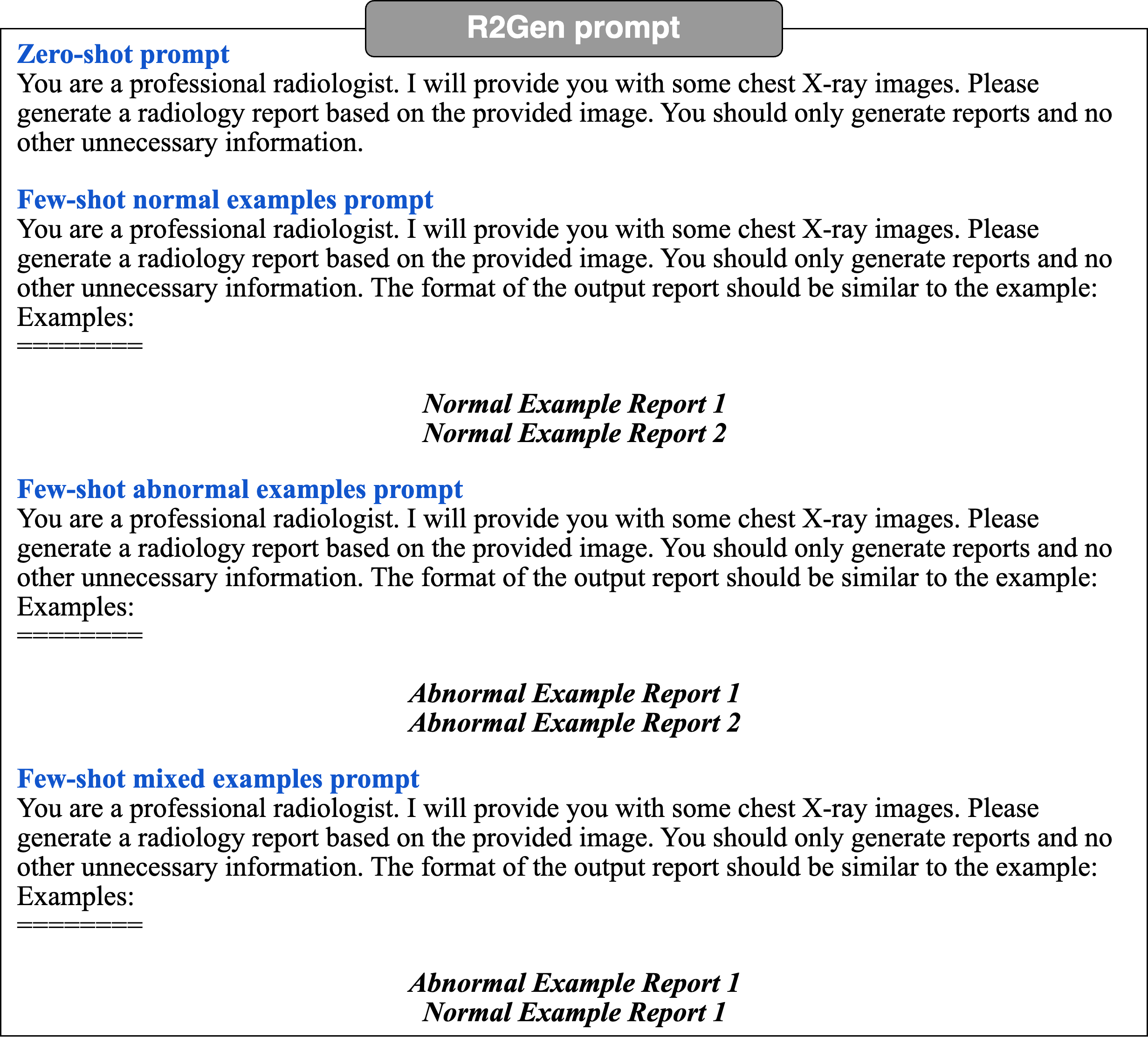}
\end{center}
\caption{R2Gen Prompt Examples. Three types of prompt settings are tested, including one zero-shot prompt and three few-shot prompts.}
\label{fig:RRG Prompt}
\end{figure}

\noindent Our evaluation reveals that the inclusion of both normal and abnormal example reports consistently results in higher-quality report generation. More discussions about this finding are provided in our case study. Also, details regarding our selected example reports are given in~\ref{sec:Few-shot prompt}. Subsequently, we employ the \textbf{few-shot mixed-example prompt} to evaluate GPT-4V on the MIMIC-CXR benchmark~\citep{2019MIMIC}. Our primary focus lies on the \textbf{zero-shot} and \textbf{few-shot} scenarios, deliberately avoiding complex techniques such as chain-of-thought~\citep{wei2022chain} or ensembling strategies~\citep{wang2022self} to manage the complexity of our efforts. \\ \\ 
\noindent \textbf{Evaluation Metrics:} In our evaluation of the generated reports, we calculate a range of metrics including BLEU, ROUGE, METEOR, and CIDEr. These metrics, widely recognized in the field of natural language processing (NLP), offer diverse perspectives on the quality of the generated text. BLEU measures the similarity of the generated text to a set of reference texts, focusing on the precision of word choices. ROUGE assesses the recall aspect, evaluating how much of the reference content is captured in the generated text. METEOR considers both precision and recall, accounting for synonymy and paraphrasing. CIDEr, originally developed for image captioning, evaluates the similarity of the generated text to multiple reference texts, emphasizing the consensus among them. Following the computation of these scores, we conduct human evaluations to ascertain the consistency between these computational assessments and human judgment. The human evaluation involves reviewers analysing the quality of the generated reports in terms of relevance, accuracy, coherence, and overall appropriateness. This combined approach of using both NLP metrics and human judgment is crucial for a more accurate evaluation. While metrics like BLEU, ROUGE, METEOR, and CIDEr offer quantitative insights into specific aspects of text quality, human evaluation provides a qualitative assessment, capturing subtleties and nuances that NLP metrics may overlook. By comparing the results from both evaluation methods, we validate the effectiveness of the generated reports from multiple perspectives, ensuring a comprehensive and reliable assessment of their quality. This process helps in determining whether the generated reports are not only linguistically correct but also contextually and semantically appropriate, aligning with human understanding and expectations.

\subsubsection {Medical Visual Question Answering}
\begin{figure}[h]
\centering
\begin{center}
%\framebox[4.0in]{$\;$}
\includegraphics[width=0.45\textwidth]{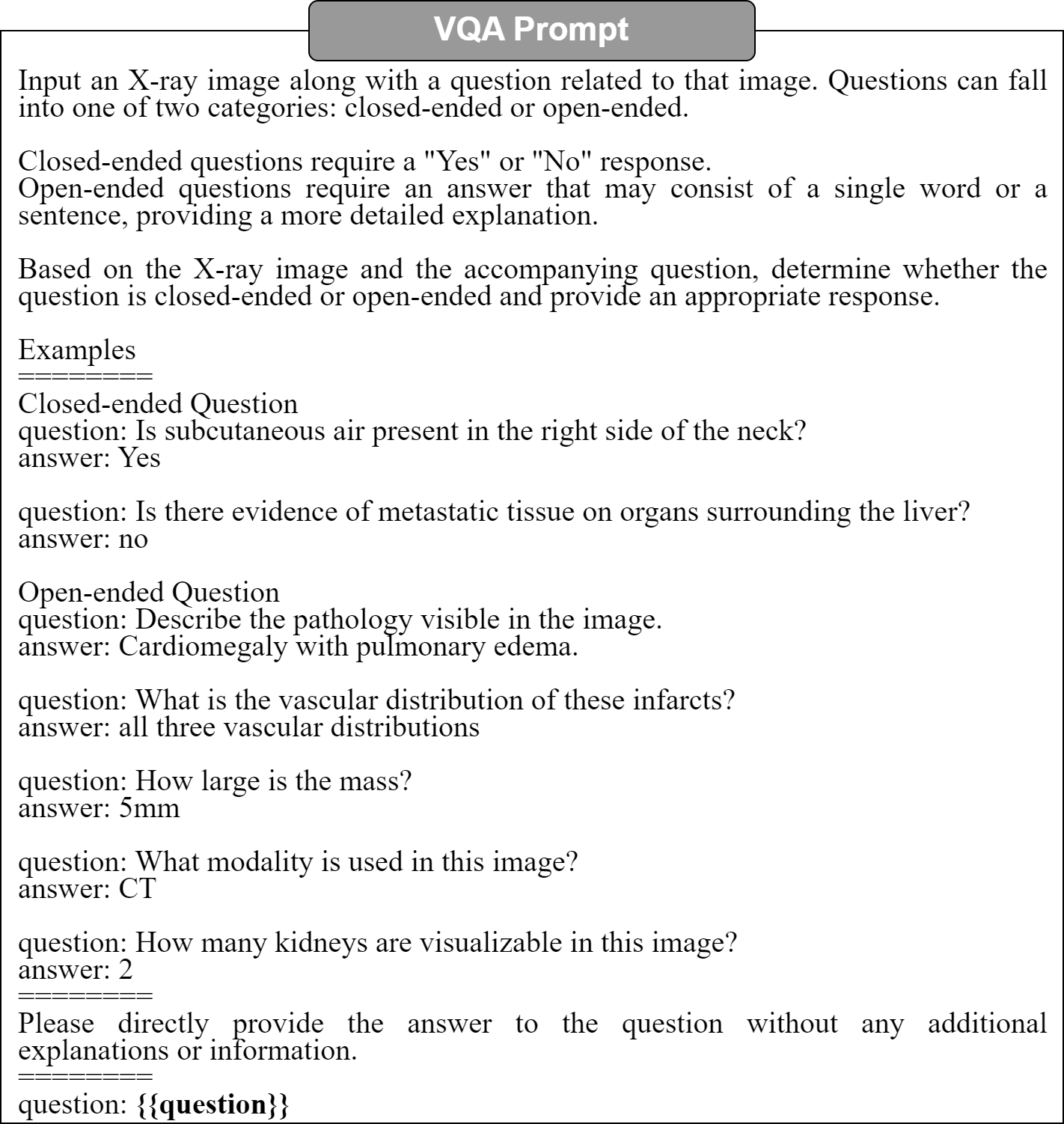}
\end{center}
\caption{VQA Prompt Example. The content between the double braces {{}} is replaced with specific questions.}
\label{fig:VQA Prompt Method}
\end{figure}

\noindent For medical VQA, we employ the ``few-shot prompt" strategy to address a significant limitation associated with the existing VQA datasets, namely the limited range of their predetermined answer set. To enhance the alignment of the answers generated by GPT-4V with the dataset, our strategy involves introducing GPT-4V to the concept of categorizing VQA questions based on the expected answer types. This approach enables the model to differentiate between close-end questions typically requiring short and concise answers (e.g., ``yes" or ``no"), and open-end questions demanding more detailed responses. By training GPT-4V with this categorization approach, we enhance GPT-4V's ability to adapt the length of its responses appropriately. For close-end questions, it learns to provide brief answers, while for open-end ones, it comprehends the need for more elaborate responses. This nuanced response mechanism significantly improves the relevance and accuracy of GPT-4V's answers, thereby increasing the effectiveness of our evaluations.\\ \\
\noindent \textbf{VQA Prompt:} Our VQA prompt follows the template in Figure~\ref{fig:VQA Prompt Method}. We provide seven
examples to guide the model in generating responses consistent with the dataset’s format. Without these examples, GPT-4V tends to produce more unconstrained answer text, complicating the task of comparing the predicted answers with the ground truth.\\

\noindent \textbf{Evaluation Metrics:} To analyse GPT-4V's performance in medical VQA, we employ different evaluation metrics for close-end and open-questions, respectively. For close-end questions, we calculate the model's prediction accuracy and compare its performance with that of the state-of-the-art (SOTA) VQA methods which are mainly classification-based approaches. This comparison is crucial to understanding how GPT-4V ranks in the current landscape of VQA technologies, especially those tailored for medical applications.
For open-end questions, we calculate BLEU scores (specifically, BLEU-4), a widely used metric in NLP to assess the quality of machine-generated text.
This choice is based on the nature of open-end questions which typically require more detailed and varied responses than close-end questions. Unlike close-end questions with straightforward single-word answers, open-end questions permit a range of possible correct answers, each potentially phrased differently.
After calculating the BLEU scores for open-end questions, we proceed to human evaluation to assess the consistency between the BLEU scores and human judgment. This step is crucial for a more effective evaluation of the accuracy of the generated answers.
Human evaluation involves a trained radiologist reviewing and rating the quality of the model-generated answers. This method is particularly valuable as BLEU scores, while offering a quantitative measure of text similarity, might not fully capture the nuances of meaning, relevance, and context that human evaluators can discern. By comparing the outcomes of the BLEU score analysis with human assessments, we gain a more holistic understanding of the GPT-4V's performance.
This dual approach of combining BLEU scores with human evaluation allows us to cross-verify the effectiveness of the model's answers.

\subsubsection {Medical Visual Grounding:}

\begin{figure}[h]
\begin{center}
%\framebox[4.0in]{$\;$}
\includegraphics[width=0.5\textwidth]{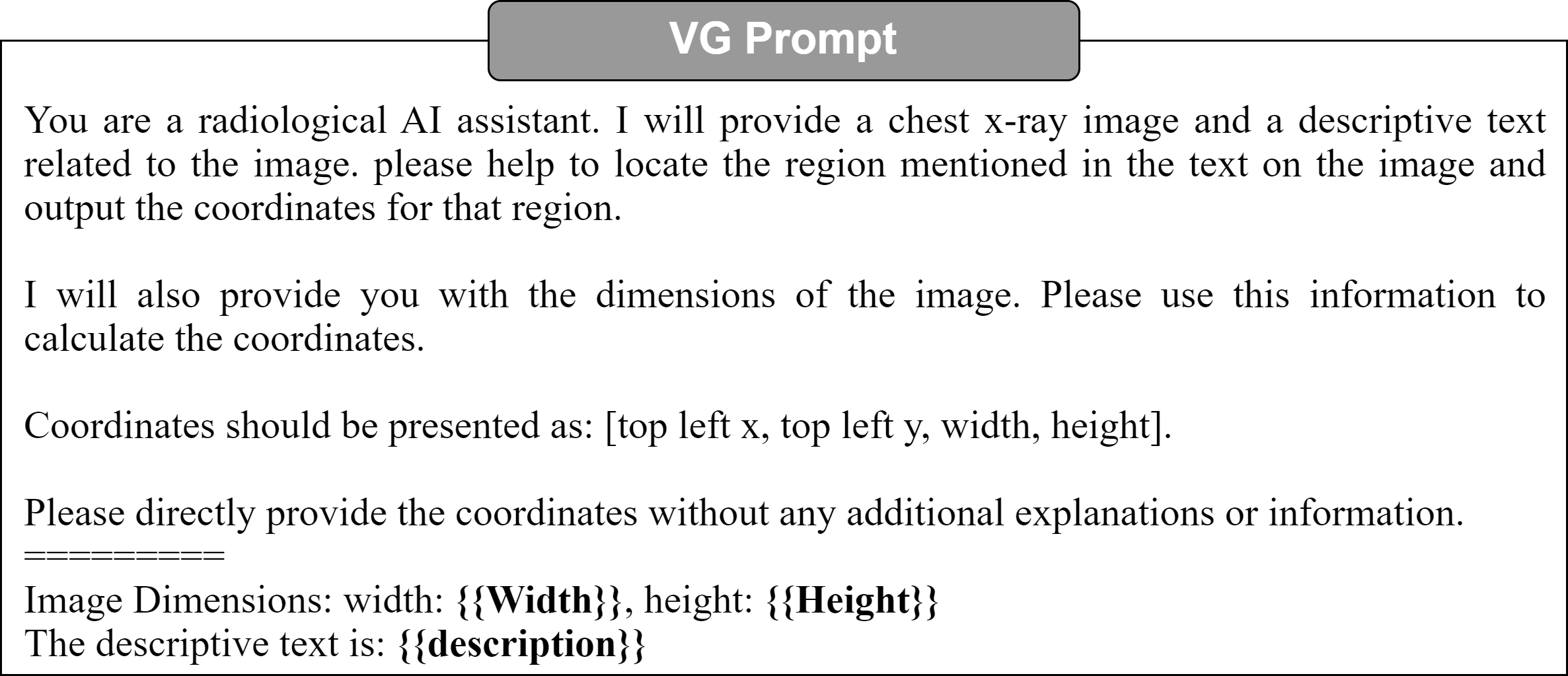}
\end{center}
\caption{VG Prompt Example. The content between double braces {{}} is replaced with specific image width, height, and description text related to the image.}
\label{fig:VG Prompt Method}
\end{figure}

For the Medical Visual Grounding (VG) task, we employ a straightforward prompt without incorporating examples. We opt for this approach to prevent potential restrictions on GPT-4V's response to only those scenarios covered in the examples, thus preserving its ability to generalize. Our goal is to avoid steering the model towards producing results solely based on the provided examples. In our prompt design, we focus solely on informing GPT-4V about the required output format, i.e., the coordinates of the bounding box. This approach guides the model in understanding our specific requirements, particularly in generating bounding boxes for visual grounding tasks.
\\ \\
\noindent \textbf{Visual Grounding Prompt:} 
We design a specific type of prompt that significantly improves the model's understanding and its ability to accurately generate bounding boxes. The prompt is illustrated in Figure~\ref{fig:VG Prompt Method}. \\

\noindent \textbf{Evaluation Metrics:} We evaluate by calculating the mean Intersection over Union (mIoU) across all samples in the benchmark dataset. The mIoU metric is a standard measure used in the field to assess the accuracy of object localization and segmentation in images. By computing the mIoU, we could gauge how precisely the GPT-4V identifies and outlines relevant areas in the images compared to the ground truth.
After calculating the mIoU, we conduct a comparative analysis with the SOTA VG methods. This comparison allows us to benchmark GPT-4V's performance against the currently leading methods in the field. By comparing the mIoU results of GPT-4V with those of the SOTA methods, we could gain a clear understanding of the strengths and weaknesses of GPT-4V in the context of the broader research landscape.

\subsection{Evaluation Datasets}
\noindent \textbf{Radiology Report Generation Dataset: MIMIC-CXR}~\citep{2019MIMIC}, the largest publicly available dataset in this domain, includes both chest radiographs and unstructured textual reports. This dataset comprises a total of 377,110 chest X-ray images and 227,835 corresponding reports, obtained from 64,588 patients who underwent examinations at the Beth Israel Deaconess Medical Center between 2011 and 2016. To facilitate fair and consistent comparisons, we follow the official partitioning provided by MIMIC-CXR, resulting in a test set containing 3,858 samples. For human evaluation, 100 report pairs comprising both the ground truth and their corresponding reports generated by GPT-4V are randomly selected from the MIMIC-CXR dataset and rated by a trained radiologist.\\ 

\noindent \textbf{VQA Dataset: VQA-RAD}~\citep{lau2018dataset} is one of the most widely utilised radiology datasets. It comprises 315 images along with 3515 question-answer pairs, ensuring that each image corresponds to at least one question-answer pair. The questions encompass 11 distinct categories, including ``anomalies'', ``properties'', ``color'', ``number'', ``morphology", ``organ type", ``other", and ``section". A noteworthy 58\% of these questions are designed as close-end queries, while the remainder takes the form of open-end inquiries. These images predominantly feature the head, chest, and abdomen regions of the human body. The dataset is officially partitioned into a training subset of 3064 question-answer pairs and a test subset of 451 question-answer pairs for evaluation. For human evaluation, 100 open-end question-answer pairs are randomly selected from the VQA-RAD dataset and marked by a trained radiologist.\\

\noindent \textbf{Visual Grounding Dataset: MS-CXR}~\citep{boecking2022making} dataset is a valuable resource for biomedical vision-language processing, featuring 1162 image-sentence pairs with bounding boxes and corresponding phrases. It was meticulously annotated by board-certified radiologists, covering eight cardiopulmonary radiological findings, each having an approximately equal number of pairs. This dataset offers both reviewed and edited bounding boxes/phrases and manually created bounding box labels from scratch. What sets MS-CXR apart is its focus on complex semantic modeling and real-world language understanding, challenging models with joint image-text reasoning and tasks like parsing domain-specific location references, complex negations, and variations in reporting style. It serves as a benchmark for phrase grounding and has been instrumental in demonstrating the effectiveness of principled textual semantic modeling for enhancing self-supervised vision-language processing.

%\noindent \textbf{Human Evaluation Dataset}
%This small dataset was evaluated by a radiologist. We randomly chose 100 pairs of reports, consisting of ground truth reports and those generated by GPT-4V, and had them evaluated by medical professionals who classified them into three categories: high, medium, and low. These grades signify that 'high' is largely consistent, 'medium' correctly identifies at least one lesion, and 'low' is completely irrelevant.

\subsection{Evaluation Result}
\subsubsection{Quantitative Results}

\noindent\textbf{Radiology Report Generation}
% \noindent \textbf{Dataset: MIMIC-CXR} ~\citep{2019MIMIC}, the largest publicly available dataset in this domain, includes both chest radiographs and unstructured textual reports. This dataset comprises a total of 377,110 chest X-ray images and 227,835 corresponding reports, obtained from 64,588 patients who underwent examinations at the Beth Israel Deaconess Medical Center between 2011 and 2016. To facilitate fair and consistent comparisons, we followed the official partitioning provided by MIMIC-CXR, resulting in a test set containing 3,858 samples.

\begin{figure}[h]
\begin{center}
%\framebox[4.0in]{$\;$}
\includegraphics[width=0.45\textwidth]{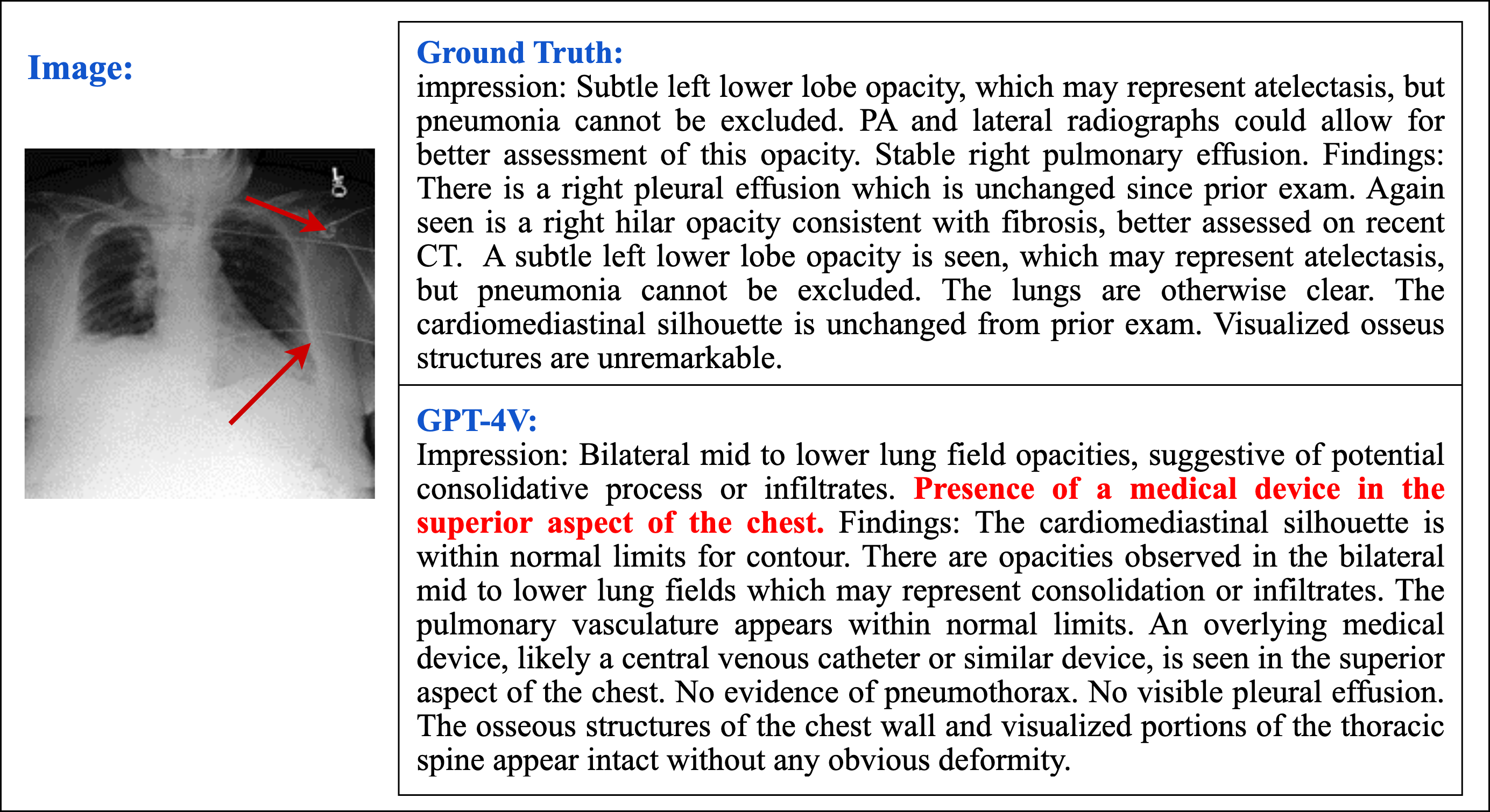}
\end{center}
\caption{An illustration of GPT-4V's capability in generating medical reports using our designed few-shot mixed-example prompt. The ground truth does not mention a medical device but one is visibly present in the image, marked by red arrows. GPT-4V demonstrates the ability to recognize and describe the medical device in the generated report.}
\label{fig: Beyond Ground Truth}
\end{figure}

\underline{Comparison with SOTA Methods:} Table~\ref{Table:ComparisonWithSOTA_RPG} presents a performance comparison between the GPT-4V model and SOTA methods using the MIMIC-CXR dataset \citep{2019MIMIC}. The comparison methods encompass standard image captioning techniques, including Show-Tell~\citep{vinyals2015showandtell}, Att2in~\citep{xu2016showattandtell}, AdaAtt~\citep{2017Knowing}, Transformer~\citep{vaswani2017attentionisallyouneed}, and M2Transformer~\citep{cornia2020meshedmemory}. Additionally, we compare with the radiology report generation methods, specifically R2Gen~\citep{chen2020generating}, R2GenCMN~\cite{chen2022cross}, MSAT~\citep{wang2022medical}, and METransformer~\citep{wang2023metransformer}. As aforementioned, we employ few-shot mixed-example prompts to help GPT-4V generate medical reports.
\begin{table*}[h]
\centering
\caption{Comparison on the MIMIC-CXR dataset.}
\vspace{1mm}
\label{Table:ComparisonWithSOTA_RPG}
\fontsize{8pt}{11pt}\selectfont
\begin{tblr}{
  colsep = 3pt,
  column{1} = {l},
  column{2-8} = {c},
  vline{2} = {-}{},
  hline{1-2} = {-}{},
  hline{13} = {1-12}{},
}

Methods        & BLEU-1 & BLEU-2 & BLEU-3 & BLEU-4 & ROUGE & METEOR & CIDEr \\
Show-Tell~\citep{vinyals2015showandtell}     & 0.308  & 0.190  & 0.125  & 0.088  & 0.256 & 0.122  & 0.096 \\
Att2in~\citep{xu2016showattandtell}        & 0.314  & 0.198  & 0.133  & 0.095  & 0.264 & 0.122  & 0.106 \\
AdaAtt~\citep{2017Knowing}        & 0.314  & 0.198  & 0.132  & 0.094  & 0.267 & 0.128  & 0.131 \\
Transformer~\citep{vaswani2017attentionisallyouneed}   & 0.316  & 0.199  & 0.140  & 0.092  & 0.267 & 0.129  & 0.134 \\
M2Transformer~\citep{cornia2020meshedmemory} & 0.332  & 0.210  & 0.142  & 0.101  & 0.264 & 0.134  & 0.142 \\
R2Gen~\citep{chen2020generating}         & 0.353  & 0.218  & 0.145  & 0.103  & 0.277 & 0.142  & -     \\
R2GenCMN~\citep{chen2022cross}      & 0.353  & 0.218  & 0.148  & 0.106  & 0.278 & 0.142  & -     \\
PPKED~\citep{CVPR201_PPKD}          & 0.360   & 0.224  & 0.149  & 0.106  & 0.284 & 0.149  & 0.237 \\
GSK~\citep{2021Knowledge}            & 0.363  & 0.228  & 0.156  & 0.115  & 0.284 & -      & 0.203 \\
MSAT~\citep{wang2022medical}           & 0.373  & 0.235  & 0.162  & 0.120  & 0.282 & 0.143  & 0.299 \\
METransformer~\citep{wang2023metransformer}  & \textbf{0.386}  & \textbf{0.250}  & \textbf{0.169}  & \textbf{0.124}  & \textbf{0.291} & \textbf{0.152}  & \textbf{0.362} \\
GPT-4V~\citep{OpenAI2023GPT4TR}        &0.338  &0.190  &0.109  &0.061  &0.240  &0.125   &0.033  \\ \hline    
\end{tblr}
\end{table*}
From Table~\ref{Table:ComparisonWithSOTA_RPG}, it is clear that radiology report generation models such as METransformer, MSAT, and R2Gen exhibit top-tier performance. Nevertheless, GPT-4V's capability to generate medical reports is impressive, considering it is designed as a general-purpose model. Leveraging the advantages of an extensive dataset for pretraining, GPT-4V performs well in several metrics, including BLEU~\citep{Kishore2002bleu}, ROUGE~\citep{lin-2004-rouge}, and METEOR~\citep{banerjee-lavie-2005-meteor}. However, when compared to models specifically trained on MIMIC-CXR, GPT-4V exhibits a performance gap, particularly evident in  CIDEr~\citep{vedantam2015cider}. This discrepancy arises because the CIDEr metric differently scores words based on their occurrence frequencies, potentially affecting GPT-4V's performance when it fails to 
generate certain MIMIC-CXR-specific words, yielding relatively lower scores. Our evaluation reveals that GPT-4V possesses the capacity to generate information not present in the ground truth but visually evident in the image. This phenomenon contributes to GPT-4V's relatively lower performance on metrics such as BLEU which primarily assesses word-match rates. One example is shown in Figure~\ref{fig: Beyond Ground Truth}.

\underline{Clinical Efficacy on MIMIC-CXR Dataset:} We assess the clinical efficacy of the GPT-4V model on the MIMIC-CXR dataset, which achieves a precision of 0.353, recall of 0.365, and F1 score of 0.330 in Table~\ref{Table:ComparisonWithSOTA_RPG}. Numerically, these results are not too bad compared to the two high-performing models METransformer~\citep{wang2023metransformer} and R2GenGPT~\citep{wang2023r2gengpt}. GPT-4V demonstrates competitive clinical efficacy. Notably, the gap between GPT-4V and other models in terms of clinical efficacy is relatively smaller than that observed in traditional NLP metrics which primarily measure lexical overlap with ground truth reports. This suggests that, objectively, GPT-4V exhibits an impressive capability in radiology report generation, emphasizing its potential clinical accuracy and practicality.

\begin{table}[h]
\centering
\caption{Evaluation of Clinical Efficacy on MIMIC-CXR dataset.}
\vspace{1mm}
\label{Table:ComparisonWithSOTA_RPG}
\fontsize{8pt}{11pt}\selectfont
\begin{tblr}{
  colsep = 3pt,
  column{1} = {l},
  column{2-4} = {c},
  vline{2} = {-}{},
  hline{1-2} = {-}{},
  hline{4} = {-}{},
}

Models        & Precision & Recall & F1    \\
METransformer~\citep{wang2023metransformer} & 0.364 & 0.309 & 0.334 \\
R2GenGPT~\citep{wang2023r2gengpt}      &  \textbf{0.392} & \textbf{0.387} & \textbf{0.389} \\
GPT-4V        & 0.353 & 0.365 & 0.330  \\ \hline    
\end{tblr}
\end{table}
\noindent \textbf{Medical Visual Question Answering} 

%\underline{Comparison with SOTA Methods:} 
\noindent Table~\ref{Results on VQA-RAD benchmark} shows a performance comparison between the GPT-4V model and SOTA VQA methods using the VQA-RAD dataset. The compared methods include StAn~\citep{he2020pathvqa}, BiAn~\citep{he2020pathvqa}, MAML~\citep{finn2017model}, MEVF~\citep{nguyen2019overcoming}, MMQ~\citep{do2021multiple}, PubMedCLIP~\citep{eslami2021does}, MMBERT~\citep{khare2021mmbert}, and the Q2ATransformer~\citep{liu2023q2atransformer}. Recall that we report accuracies for close-end questions and BLEU scores for open-end questions. %\textbf{\textit{Our evaluation primarily uses accuracy as the metric and concentrates on close-end questions that have single-word answers, such as ``yes" or ``no". We chose not to include open-end questions in this comparison because accuracy isn't an effective measure in this context. This is due to a fundamental difference in approaches: while other methods utilise classification techniques, GPT-4V employs a generative approach. In the case of open-end questions, there are seldom exact matches with the predefined answer pool, making accuracy a less reliable indicator. For the open-end questions, we instead measured performance using the BLEU score, a common metric for evaluating the quality of text generation. The BLEU score for these questions was 0.1155, indicating the challenges in generating precise answers in an open-end format within the medical domain. We conduct the following human evaluation to evaluate the quality of open-end questions, which proved that the open-end question result is not so bad. }}
From Table~\ref{Results on VQA-RAD benchmark}, it can be seen that GPT-4V achieves an accuracy of 61.4\% for the close-end questions, notably lower than other published results. In our experiment, we meticulously define open-end and close-end questions for GPT-4V in our prompt design, intending to guide its response generation effectively. Despite this, we observe that GPT-4V sometimes produces relatively long answers even for close-end questions that typically require brief single-word responses, such as ``yes" or ``no". This deviation from the expected answer format leads to a lower accuracy score for these instances. Additionally, GPT-4V's BLEU score for open-end questions is also modest at 0.116. This can be attributed to GPT-4V's robust text generation capacity, leading to a greater diversity in its outputs. For example, GPT-4V can represent the same object in multiple ways leading to varied responses. This diversity suggests that traditional evaluation metrics like BLEU, which are based primarily on the overlap of words and lack of a deep understanding of textual meaning, are insufficient to assess the quality of GPT-4V's varied answers. %BLEU does not deeply understand the meaning of the text; it merely quantifies similarity in word usage. 
Given this limitation, we complement our evaluation with human assessments in Section~\ref{Human Evaluation Results} to provide a more appropriate and comprehensive evaluation. Human evaluation can capture subtleties and interpretations that purely text-based metrics like BLEU may overlook, especially when GPT-4V's responses exhibit diversity and nuance beyond the scope of traditional metrics. \\

% \begin{table}
% \centering
% \caption{Results on VQA-RAD benchmark}
% \label{Results on VQA-RAD benchmark}
% \scalebox{0.7}{
% \begin{tblr}{
%   row{odd} = {c},
%   row{4,6,8,10,12} = {c},
%   cell{2}{1} = {r=12}{},
%   cell{2}{2-4} = {c},
%   cell{4,6,8}{2} = {r=2}{},
%   vline{2-5} = {1-13}{},
%   hline{1-2,14} = {-}{},
%   hline{3,5,7,9} = {3-4}{},
%   hline{3,4,6,8,10-13} = {2-4}{},
% }
% Dataset & Reference Methods & Fusion Method & Close-end \\
% VQA-RAD & StAn~\citep{he2020pathvqa}              & SAN           & 57.2      \\
%         & BiAn~\citep{he2020pathvqa}              & BAN           & 67.9      \\
%         & MAML~\citep{finn2017model}              & SAN           & 69.7      \\
%         &                   & BAN           & 72.4      \\
%         & MEVF~\citep{nguyen2019overcoming}              & SAN           & 74.1      \\
%         &                   & BAN           & 75.1      \\
%         & MMQ~\citep{do2021multiple}               & SAN           & 75.7      \\
%         &                   & BAN           & 75.8      \\
%         & PubMedCLIP~\citep{eslami2021does}        & -             & 80        \\
%         & MMBERT~\citep{khare2021mmbert}           & -             & 77.9      \\
%         & Q2ATransformer~\citep{liu2023q2atransformer}    & -             & 81.2      \\
%         & GPT-4V~\citep{OpenAI2023GPT4TR}            & -             & 61.40     
% \end{tblr}
% }
% \end{table}

\begin{table}
\centering
\caption{Results on VQA-RAD benchmark}
\label{Results on VQA-RAD benchmark}
\scalebox{0.7}{
\begin{tblr}{
  %row{odd} = {c},
  %row{4,6,8,10,12} = {c},
  %cell{2}{1-4} = {c},
  column{2-4} = {c},
  cell{4,6,8}{1} = {r=2}{},
  vline{2-4} = {1-13}{},
  hline{1-2,14} = {-}{},
  hline{3,5,7,9} = {1-4}{},
  hline{3,4,6,8,10-13} = {1-4}{},
  hline{13} = {1-4}{},
}
Reference Methods & Fusion Method & \thead{Accuracy (\%) \\ (Close-end)} &\thead{BLEU-4 \\(Open-end)} \\
StAn~\citep{he2020pathvqa}              & SAN           & 57.2     & - \\
BiAn~\citep{he2020pathvqa}              & BAN           & 67.9      & - \\
MAML~\citep{finn2017model}              & SAN           & 69.7      & - \\
                   & BAN           & 72.4      & - \\
MEVF~\citep{nguyen2019overcoming}              & SAN           & 74.1     & - \\
                   & BAN           & 75.1     & - \\
MMQ~\citep{do2021multiple}               & SAN           & 75.7     & - \\
                   & BAN           & 75.8    & -  \\
PubMedCLIP~\citep{eslami2021does}        & -             & 80       & - \\
MMBERT~\citep{khare2021mmbert}           & -             & 77.9     & - \\
Q2ATransformer~\citep{liu2023q2atransformer}    & -             & 81.2   & -   \\
GPT-4V~\citep{OpenAI2023GPT4TR}            & -             & 61.40     & 0.1155
\end{tblr}
}
\end{table}

%--------------%
\noindent \textbf{Medical Visual Grounding}

%\underline{Comparison with SOTA Methods:} 
Table~\ref{VGComparisonWithSOTA} reports a comparative performance of GPT-4V against various SOTA visual grounding methods using the MS-CXR dataset. We compare GPT-4V with  a range of advanced methods in the field: BioViL~\citep{boecking2022making}, BioViL-T~\citep{bannur2023learning}, RefTR~\citep{li2021referring}, VGTR~\citep{du2022visual}, SeqTR~\citep{zhu2022seqtr}, TransVG~\citep{deng2021transvg}, and MedRPG~\citep{chen2023medical}. Each of these models represents a significant approach or innovation in medical image analysis, making them suitable benchmarks for evaluating GPT-4V's performance. By using the mIoU metric, we can quantitatively assess how well GPT-4V and other models perform in terms of accurately identifying and delineating relevant patterns within the medical images of the MS-CXR dataset. GPT-4V's performance on the MS-CXR dataset yields an mIoU of 0.083, significantly lower than all published benchmarks. While GPT-4V demonstrates a level of comprehension in visual grounding, it struggles to accurately identify medical organs and pathological signs, leading to imprecise bounding box predictions. It should be noted that the recent SoM~\citep{yang2023set} model has shown substantial improvements in this area by segmenting and labeling images before grounding, enhancing performance on generic images. However, its effectiveness on medical images demanding finer details remains untested. Further research is necessary to assess its applicability to medical imaging.
\begin{table}[h]
\centering
\caption{mIoU(\%) results on MS-CXR benchmark.}
\vspace{1mm}
\label{VGComparisonWithSOTA}
\fontsize{8pt}{11pt}\selectfont
\begin{tblr}{
  colsep = 3pt,
  column{1} = {l},
  column{2} = {c},
  vline{2} = {-}{},
  hline{1-2} = {-}{},
  hline{9} = {1-9}{},
}
Methods        & mIoU(\%)  \\
BioViL~\citep{boecking2022making}     & 22.9   \\
BioViL-T~\citep{bannur2023learning}              & 24.3             \\
RefTR~\citep{li2021referring}               & 50.11             \\
VGTR~\citep{du2022visual}             & 53.58               \\
SeqTR~\citep{zhu2022seqtr}       & 56.63          \\
TransVG~\citep{deng2021transvg}               & 58.91            \\
MedRPG~\citep{chen2023medical}       & \textbf{59.37}               \\
GPT-4V~\citep{OpenAI2023GPT4TR}            & 8.33 \\ \hline    
\end{tblr}
\end{table}

\subsubsection{Human Evaluation Results}
\label{Human Evaluation Results}
\noindent This section details our approach to evaluating and analysing human assessments of radiology reports and VQA tasks. Specifically, we choose not to include human evaluation for the visual grounding task. The reason is that the output of visual grounding is typically a simple bounding box, rather than text. The quality of these bounding boxes can be straightforwardly assessed through visualization, making human evaluation less critical for this aspect. However, when it comes to the evaluation of text generation, visual inspection of the words alone is insufficient for determining quality. The key aspect of evaluating the text produced by a large-scale model like GPT-4V lies in understanding the semantic meaning of the generated content. Text quality encompasses more than just the correct words. Therefore, we incorporate human evaluation as a crucial part of our assessment process for text generation. Human evaluators can provide insights into how well the model captures the intended meaning, context, and subtleties that traditional metrics may overlook. This approach allows for a more nuanced and comprehensive evaluation of the quality of text generated by GPT-4V.\\

\noindent\textbf{Radiology Reports Human Evaluation}\\
\noindent We conduct a random selection of 100 report pairs, including corresponding ground truth reports and the reports generated by GPT-4V. These pairs undergo grading by a trained radiologist, categorized into three levels: high, medium, and low, based on varying degrees of human-perceived consistency. Simultaneously, we calculate NLP scores for BLEU, ROUGE, METEOR, and CIDEr for these 100 report pairs, focusing on BLEU and CIDEr due to minimal variation in ROUGE and METEOR. 

\begin{figure*}[t!]
\centering
\includegraphics[width=\textwidth]{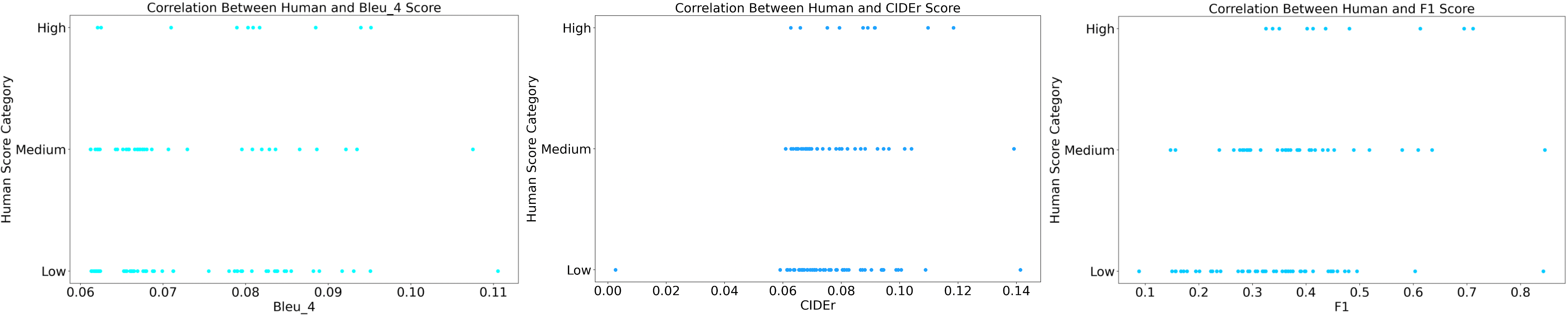}
\caption{Scatter graphs show the relationship between BLEU/CIDEr/F1 score and the human evaluation, respectively.}
\label{fig: human_consistency}
\end{figure*}

\underline{Visualization \& Distribution Analysis:} Figure~\ref{fig: human_consistency} presents scatter graphs illustrating the relationship between the BLEU/CIDEr/F1 scores and human ratings, respectively. The x-axis represents BLEU/CIDEr/F1 scores, while the y-axis depicts corresponding human scores for the same samples. Notably, a significant number of GPT-4V-generated reports with low BLEU scores ($< 0.20$) receive ``Medium" or ``High" quality evaluations in human ratings, a trend also observed with CIDEr metrics. The human rating is better aligned with the F1 score reflecting clinical efficacy, e.g., there are fewer reports with low F1 scores ($< 0.20$) rated as 'Medium' or 'High' by the human evaluator.

To align NLP scores with human ratings, we scale them to the range of [0, 100] using the formula $\hat{s}=\frac{s - \text{lowest}}{\text{highest} - \text{lowest}} \times 100$, where $s$ is the original score, and $\hat{s}$ is the scaled score. We quantize these scores into three quality ranges: low [0 ~ 40), medium [40 ~ 70), and high [70 ~ 100]. The distributions of the 100 samples are shown in Fig 6, corresponding to BLEU, CIDEr, F1, human rating, respectively. Specifically, based on the human rating, there are 10 high-, 35 medium-, and 55 low-quality reports; based BLEU scores, there are 1 high-, 2 medium-, and 97 low-quality reports; based on CIDEr, there are 5 high-, 53 medium-, and 42 low-quality reports; and based on F1 score, there 3 high-, 29 medium-, and 68 low-quality reports. Our findings indicate that traditional evaluation methods tend to yield lower scores compared to radiologist assessments, suggesting that human evaluation assigns higher scores to the reports generated by GPT-4V, implying better report quality than indicated by traditional scores.

% We randomly selected 100 report pairs (ground truth reports and GPT-4V's generated reports) and had them graded by doctors into three categories: high, medium, and low. These three grades indicate that H is largely consistent, M diagnoses at least 1 lesion correctly, and L is not relevant at all. For these 100 report pairs, we calculated their BLEU Score, ROUGE, METEOR, and CIDEr. Since ROUGE and METEOR showed minimal variation, we focused on comparing the consistency between the BLEU Score, CIDEr, and the doctors' ratings. The scatter graph depicted in Figure \ref{fig: human_consistency} illustrates the relationship between evaluation metrics like BLEU and CIDEr and human ratings. In this graph, the x-axis represents the scores of the BLEU or CIDEr, while the y-axis shows the corresponding human scores for the same samples. This visualization helps in understanding how these automated metrics correlate with human judgement in rating the samples.

% BLEU, ROUGE, METEOR, and CIDEr, borrowed from machine translation and text summarizing, are commonly used to evaluate medical image report generation. BLEU, focusing on n-gram overlap, may overlook deeper meaning and penalize brevity. METEOR, though considering semantic similarities and paraphrasing, doesn't assess coherence or overall text quality. ROUGE-L, evaluating the longest common sub-sequence, correlates well with human judgment but may miss broader quality aspects. CIDEr, measuring cosine similarity between n-gram TF-IDF representations, balances surface and semantic similarities but may be less effective for non-captioning tasks and is computationally demanding. Despite their widespread use, these metrics have limitations, reflecting the ongoing challenges in automatic radiology report generation. So, simply using those scores to evaluation the generation result generate by LLMs is not enough. So, we compared the human evaluation with traditional scores to give a more reasonable evaluation for the report generate by LLMs.

\begin{figure}[t!]
\centering
\includegraphics[width=0.4\textwidth]{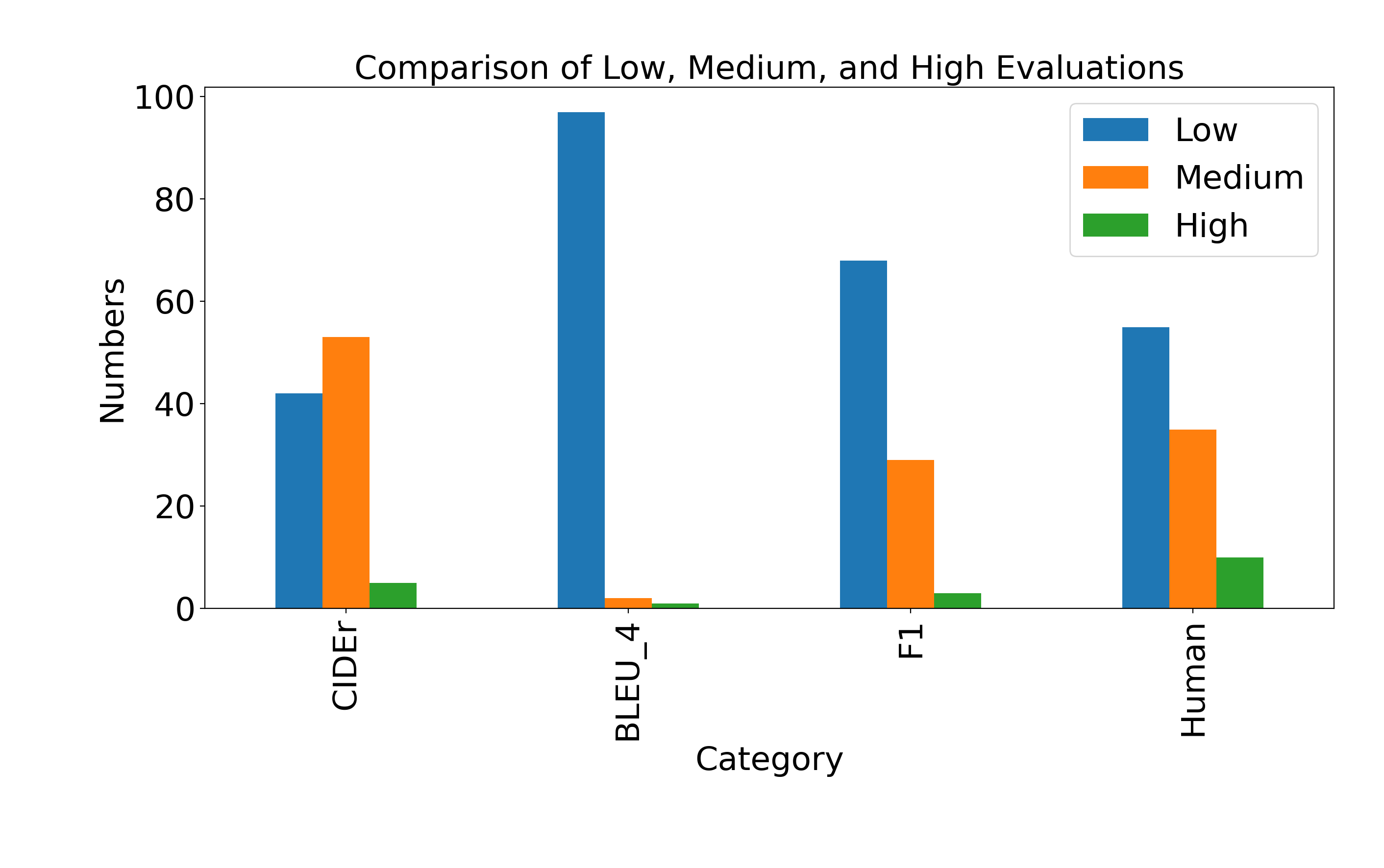}
\caption{Distributions of scores based on human evaluation, BLEU, CIDEr, and F1, respectively.}
\label{fig: distribution}
\end{figure}

\underline{Correlation \& Statistical Test}: To further assess the correlation between the NLP scores and human ratings, we calculate Kendall’s $\tau$ coefficient (i.e., Kendall rank correlation coefficient), which measures the ordinal association between two quantities. Kendall’s $\tau$ is defined as follows:
\[
\tau = \frac{\text{number of concordant pairs} - \text{number of discordant pairs}}{\text{total number of pairs} \times (\text{total number of pairs} - 1)/2}.
\]
Concordant pairs occur when both the human evaluator and the NLP metric agree on the ranking of two reports. 
%For instance, if the human evaluator ranks Report A higher than Report B, and the NLP metric also ranks Report A higher than Report B, this pair (A, B) is considered concordant. 
Conversely, discordant pairs occur when there is a disagreement between the rankings by the human evaluator and the NLP metrics. 
%For example, if the human evaluator ranks Report A higher than Report B, but the NLP metric ranks Report B higher than Report A, this pair (A, B) is discordant. 
The $\tau$ coefficient is within the range $-1 \leq \tau \leq 1$. If two random variables are independent, the expectation of $\tau$ is zero. In addition, a statistical test with the null hypothesis of $\tau = 0$ (no correlation) is conducted, and the p-value is reported. 
The p-value helps determine whether the observed agreement or disagreement in rankings between human ratings and NLP metrics is statistically significant or due to random variations.

As reported in Table 5, for the NLP scores BLEU, ROUGE, METEOR, and CIDEr, the Kendall’s $\tau$ coefficients are small values around zero, suggesting the human rating and these NLP metrics are potentially independent. This is further supported by their corresponding large p-values, all exceeding 0.1, indicating a failure to reject the null hypothesis. Thus, the correlations between human ratings and NLP metrics are not statistically significant. However, the F1 score has a small p-value of 0.002 and a moderate Kendall’s $\tau$ of 0.242, showing a statistically significant rank correlation with the human rating. In short, the statistic tests support our observations that i) the quality of GPT-4V's generated reports could be much higher than what traditional evaluation methods indicate; and ii) GPT-4V's performance in clinical efficacy aligns with human evaluation, which are higher than NLP-based metrics. On the other hand, it should also be noted that human evaluation captures more than the clinical score of F1, as the latter is still limited in detecting the occurrence of clinically relevant keywords in the generated reports without comprehending the context.

\underline{An Example of Human Evaluation}:  To further bolster the credibility of our report generation results, we present an in-depth analysis conducted by a radiologist. This expert evaluation delves into the accuracy, relevance, and overall quality of the generated reports, providing valuable professional insights. The following example illustrates a notable disparity between BLEU scores and human evaluation. It juxtaposes a ground truth report with a report generated by GPT-4V, revealing that despite a low BLEU score, the report receives a high rating from a radiologist.

\noindent\textit{Ground Truth Report: ``Impression: Bilateral parenchymal opacities right greater than left compatible with pneumonia in the proper clinical setting. Findings : There bilateral regions of consolidation at the right lung and left mid to lower lung. Findings are most concerning for bilateral infection. Moderate enlargement of the cardiac silhouette is unchanged. Multiple vascular stents are also noted. No acute osseous abnormalities. splenic calcifications are again noted."}\\

\noindent\textit{GPT-4V Predicted: ``Impression: Bilateral patchy opacities with potential for infiltrates or infectious etiology. Right-side mediastinal shift observed. Findings: The cardiomediastinal silhouette shows right-side deviation, which might be suggestive of volume loss or other pathology on the left side. Bilateral patchy opacities are noted which can represent infiltrates, raising concerns for potential infectious etiology, interstitial disease, or other inflammatory processes. The pulmonary vasculature appears within normal limits. There are no clear signs of pleural effusion or pneumothorax. The osseous structures of the thorax appear intact without evidence of acute fractures."}\\

Upon expert examination, the radiologist affirms that the predicted report accurately identifies key bilateral lung consolidation areas, describes the patchy morphology of the lesions, and provides insightful pathological inferences. The changes in the mediastinal position of the heart are also consistent with the original report. Although the direct determination of cardiac hypertrophy is not made, a positive prompt is acknowledged, aligning with chest X-ray reporting principles. Other negative diagnosis descriptions are basically consistent with the ground truth report. Consequently, the generated report receives a high score based on these professional evaluations. \\

\begin{table}[h]
\centering
\scalebox{0.75}{
\begin{tabular}{l|c|c|c|c|c}
\hline
        & Bleu-4 & ROUGE\_L & METEOR & CIDEr & F1  \\ \hline
P Value & 0.688 & 0.430 & 0.462  & 0.503 & 0.002\\
Kendall’s Tau & 0.032 & 0.063 & 0.059 & 0.053 & 0.242\\ \hline
\end{tabular}
}
\caption{Evaluation of P Value and Kendall’s Tau}
\label{f1}
\end{table}

\noindent\textbf{Visual Question Answering Human Evaluation}

\begin{figure}[t!]
\includegraphics[width=0.4\textwidth]{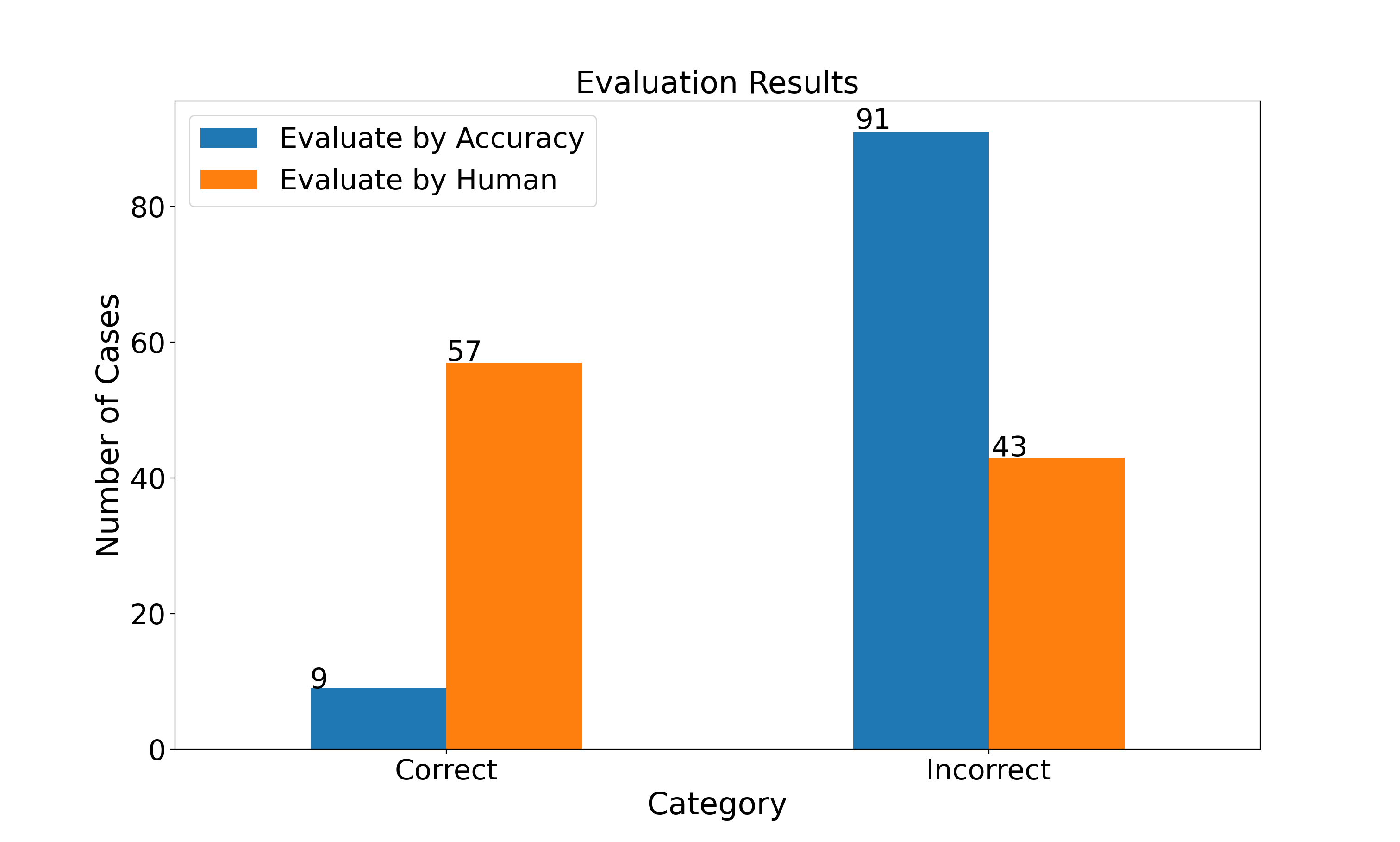}
\caption{Human evaluation of the correctness of the answers generated by GPT-4V on a  subset of VQA-RAD, in contrast to the evaluation based on the classification accuracy of the same dataset.}
\label{fig: VQA_Human_Evaluate}
\end{figure}

For this evaluation, we randomly select 100 open-end questions from VQA-RAD, together with ground truth and GPT-4V generated answers. A trained radiologist assesses the correctness of GPT-4V's answers, identifying 57 answers as correct and 43 answers as incorrect. This contrasts sharply with the traditional accuracy metric used in the predominant classification-based VQA approaches, which detects only 9 correct answers, as shown in Figure~\ref{fig: VQA_Human_Evaluate}. This disparity is rooted in the traditional classification-based VQA evaluation, which treats each predetermined answer as a class and calculates the classification accuracy. This evaluation lacks flexibility in handling the variability of answers produced by GPT-4V. Our human evaluation finds GPT-4V's answers more accurate than traditional scores, highlighting the richness of content in GPT-4V's responses not fully captured by conventional scoring methods. This underscores that the quality of GPT-4V's answers is not inferior.
%However, when evaluated by humans, the answers generated by GPT-4V were found to be much more accurate than the traditional scores suggested. This is attributed to the versatility and richness of content in GPT-4V's responses, which are not fully captured by conventional scoring methods. This also demonstrates that the quality of answers generated by GPT-4V is not inferior.
%

\begin{figure}[t!]
\includegraphics[width=0.4\textwidth]{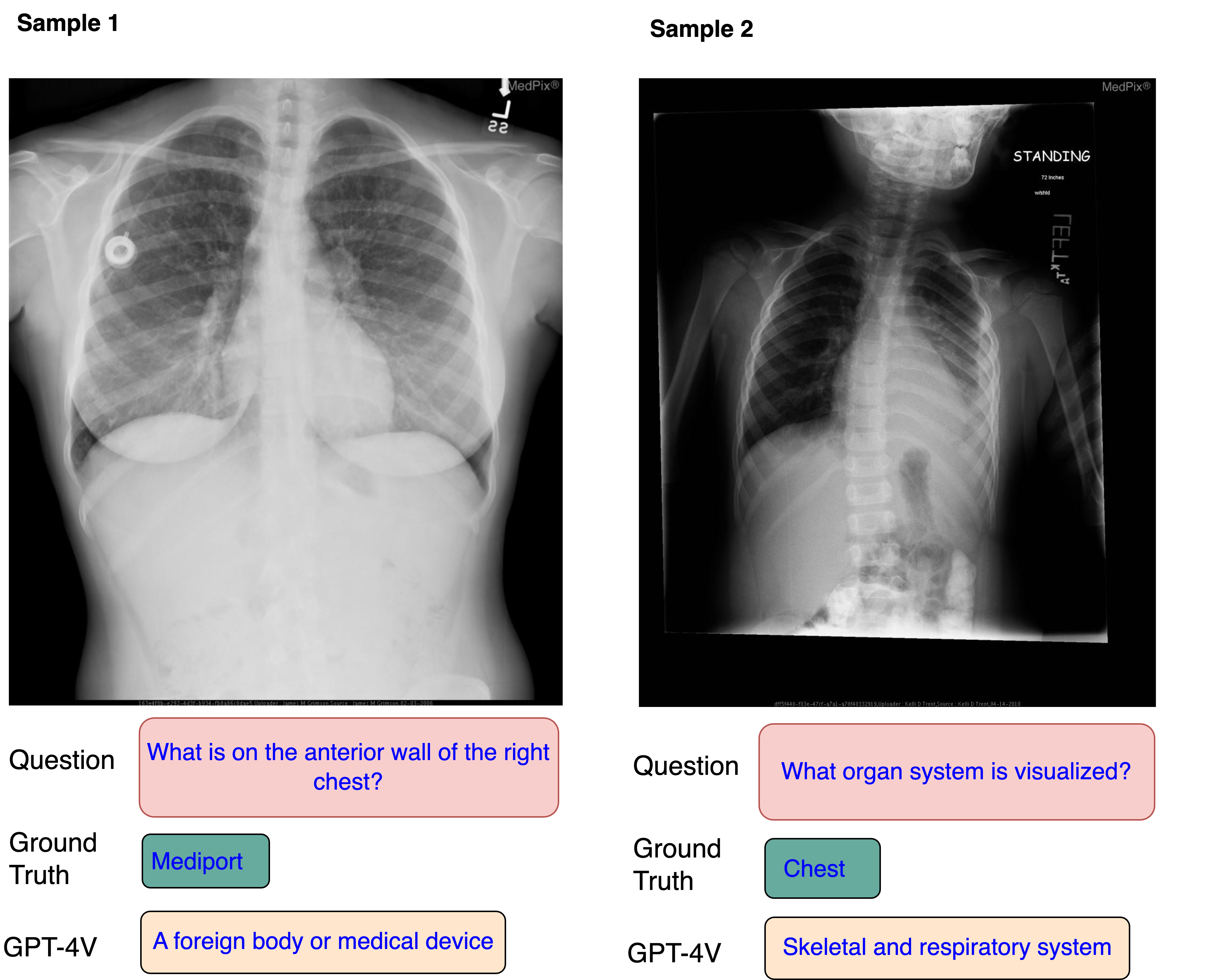}
\caption{Two examples showcase GPT-4V's performance on open-end questions.}
\label{fig: vqa_sample}
\end{figure}
Moreover, delving into the answers generated by GPT-4V, the radiologist finds GPT-4V not only matches but surpasses the accuracy of the ground truth. This superiority is attributed to the detailed and precise nature of GPT-4V's responses. Figure ~\ref{fig: vqa_sample} provides two VQA examples produced by GPT-4V.  In the right example, given the question ``what organ system is visualized", the ground truth provided only mentions the chest, which is neither rigorous nor sufficient. In contrast, GPT-4V's response includes the skeleton and respiratory system, demonstrating improved accuracy and completeness. Moreover, the answers generated by GPT-4V exhibit a higher level of generalization and comprehensiveness. As shown in the left example, GPT-4V describes ``mediport" as ``A foreign body or medical device". Additionally, GPT-4V produces answers that are more professional and readable, presenting complete sentences with medical terminology and grammatical accuracy. These findings suggest that GPT-4V holds significant learning potential, inviting further exploration through additional research.
%According to the radiologist, GPT-4V generates answers even better than the actual ground truth. This is mainly because the answers provided by GPT-4V are more detailed and accurate. For instance, GPT-4V includes additional structures such as ribs that are not present in the ground truth. Moreover, the answers generated by GPT-4V are more generalized and comprehensive. For example, ``medipot" is described as ``A foreign body or medical device". Additionally, the answers produced by GPT-4V are more professional and readable. Instead of a single word, GPT-4V generates complete sentences with medical terminology and grammatical accuracy. These highlights have surprised us and indicate that GPT-4V has enormous potential for learning, which can be further explored through more research. 

\subsubsection{Case Study}

In this part, we demonstrate GPT-4V's performance by incorporating case studies across the three specific tasks, offering a more direct and intuitive view of its capabilities.\\

\noindent\textbf{Radiology Report Generation Cases} 

\begin{figure}[h]
\centering
\begin{center}
%\framebox[4.0in]{$\;$}
\includegraphics[width=0.5\textwidth]{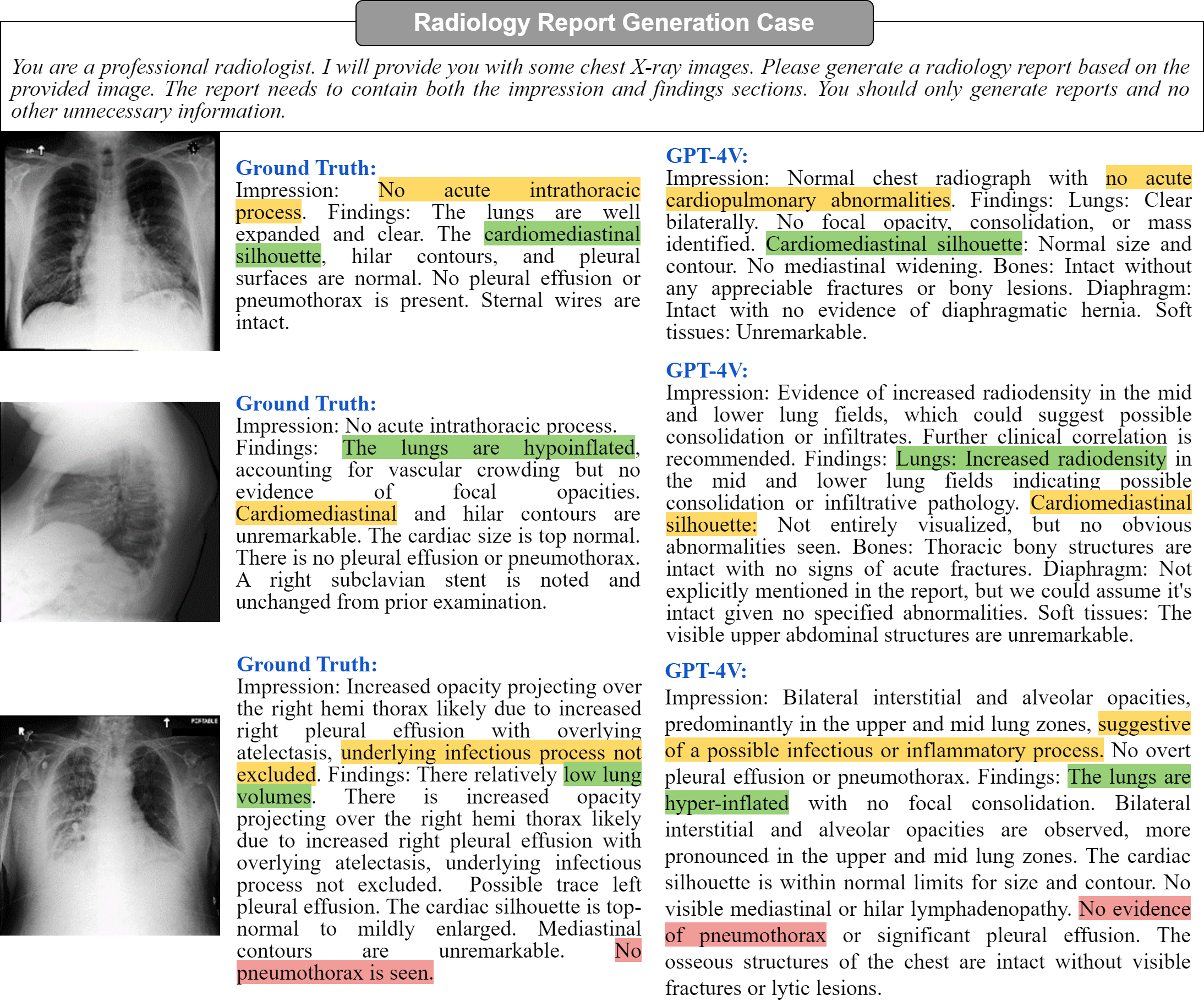}
\end{center}
\caption{R2Gen case using zero-shot prompt. GPT-4V can generate radiology reports without example reports and convey both normal and abnormal aspects. For better illustration, the key medical information in the reports is highlighted using different colours.}
\label{fig:zero_shot_case}
\end{figure}

In zero-shot scenarios, through a series of tests on multiple chest X-ray images, it is observed that GPT-4V consistently generates reports with a focus on specific anatomical organs. This phenomenon can be observed in Figure~\ref{fig:zero_shot_case}. Notably, GPT-4V tends to present information in a specific sequence, covering the information on the lung, cardiomediastinal silhouette, bones, diaphragm, and soft tissues, in the majority of the generated reports. 
% While the format of the generated reports may vary from MIMIC-CXR, the content within these reports does convey both normal and abnormal aspects of the radiographic images. 
Examining the third case in Figure~\ref{fig:zero_shot_case}, GPT-4V demonstrates its proficiency to identify both normal and abnormal aspects, e.g., \textit{``No pleural effusion or pneumothorax is present; Suggestive of a possible infectious or inflammatory process"}. These instances underscore GPT-4V's ability, even with zero-shot prompts, to generate relevant reports and identify anomalies.

\begin{figure}[h]
\centering
%\framebox[4.0in]{$\;$}
\includegraphics[width=0.5\textwidth]{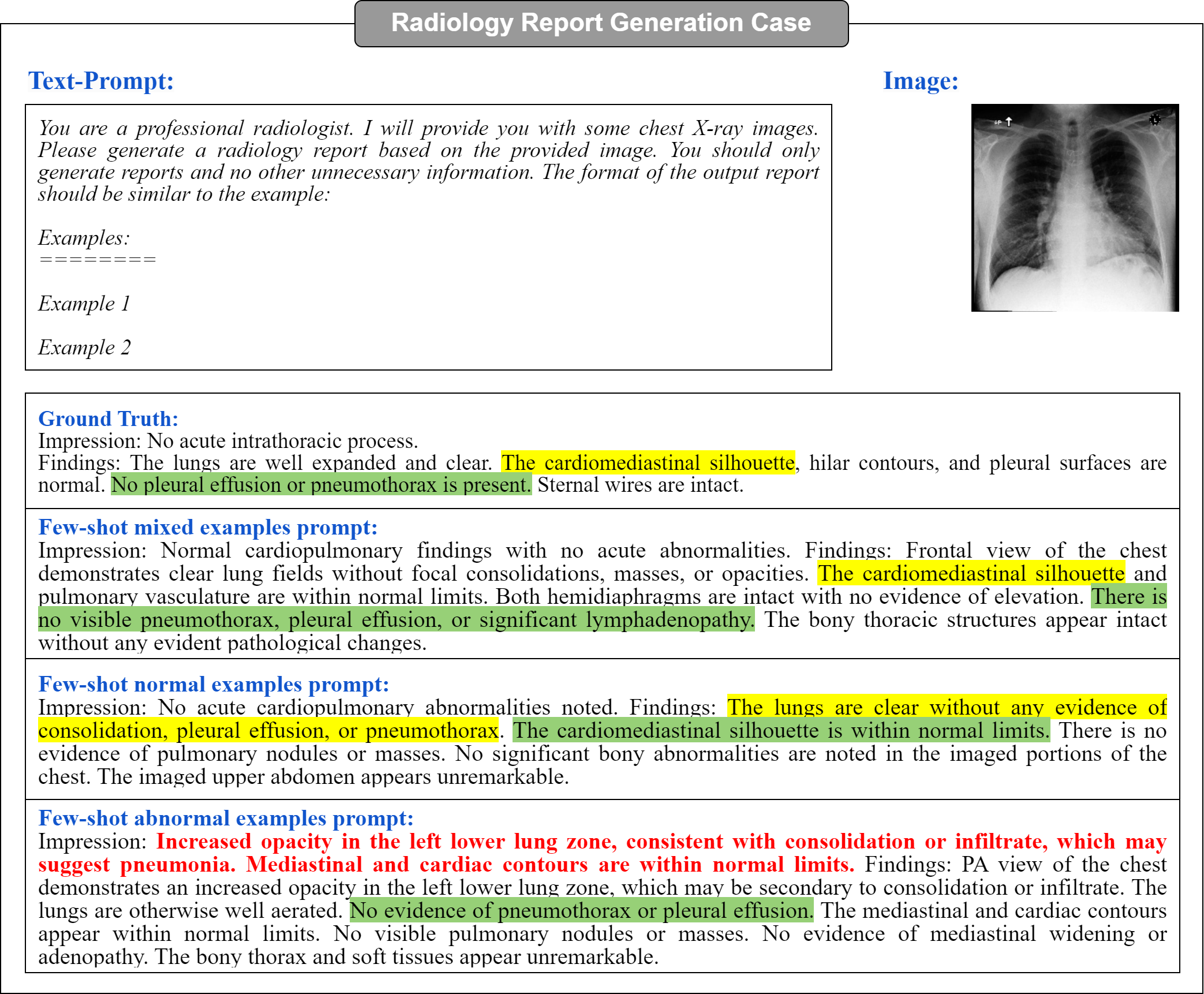}
\caption{R2Gen normal case. Key medical information in the reports is highlighted using different colors. GPT-4V is more likely to generate reports containing abnormality descriptions when the prompt consists of only abnormal examples. The text in red corresponds to descriptions of abnormal conditions.}
\label{fig:few shots normal}
\end{figure}

\begin{figure}[h]
\centering
%\framebox[4.0in]{$\;$}
\includegraphics[width=0.5\textwidth]{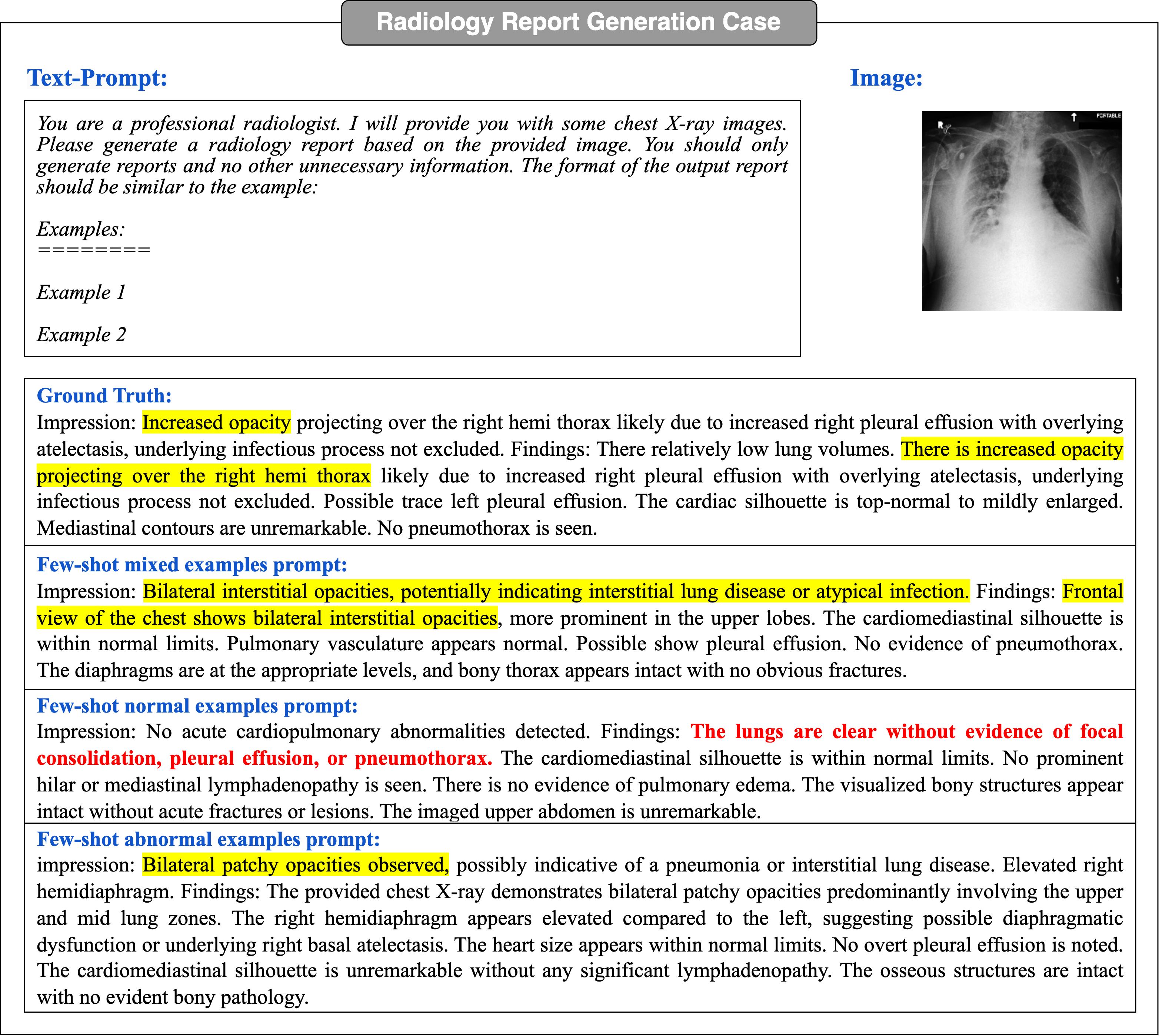}
\caption{R2Gen abnormal case. Key medical information in the reports is highlighted using different colors. GPT-4V is more likely to generate normal reports when the prompt consists of only normal examples. The text in red corresponds to descriptions of normal conditions.}
\label{fig:few shots abnormal}
\end{figure}

In few-shot scenarios, our observation indicates that different prompts significantly influence the generated reports. GPT-4V's inclination to generate normal or abnormal findings varies based on the provided example reports. Figure~\ref{fig:few shots normal} illustrates the response to a normal chest X-ray image, utilising three distinct prompt settings to guide GPT-4V in generating corresponding reports. Interestingly, it is found that the report generated either from the normal-example prompt or the mixed-example prompt describes the image as normal, which is consistent with the ground truth. In contrast, the report from the abnormal-example prompt misidentifies anomalies in the image. Meanwhile, Figure~\ref{fig:few shots abnormal} showcases an example of an abnormal chest X-ray image. It is observed that the report generated by the normal-example prompt misinterprets the lungs as normal. Our investigation emphasizes the substantial impact of the mixed-example prompt on GPT-4V's accuracy in determining the normality or abnormality of an image.

% The details of example reports in prompts are shown in Appendix~\ref{sec:Prompt Settings}. Our observations highlighted the substantial impact of prompt type on the model's output. Depending on the chosen prompt, the model displayed a clear preference either for generating normal reports or abnormal reports. 

%Notably, our subsequent tests have emphasized the substantial impact of the mixed-example prompt on GPT-4V's accuracy in determining the normality or abnormality of an image. Consequently, we used the mixed-example prompt for testing the entire MIMIC-CXR test set and in computing-related evaluation metrics.\\

\noindent\textbf{Visual Question Answering Cases} 

\begin{figure}[h]
\includegraphics[width=0.5\textwidth]{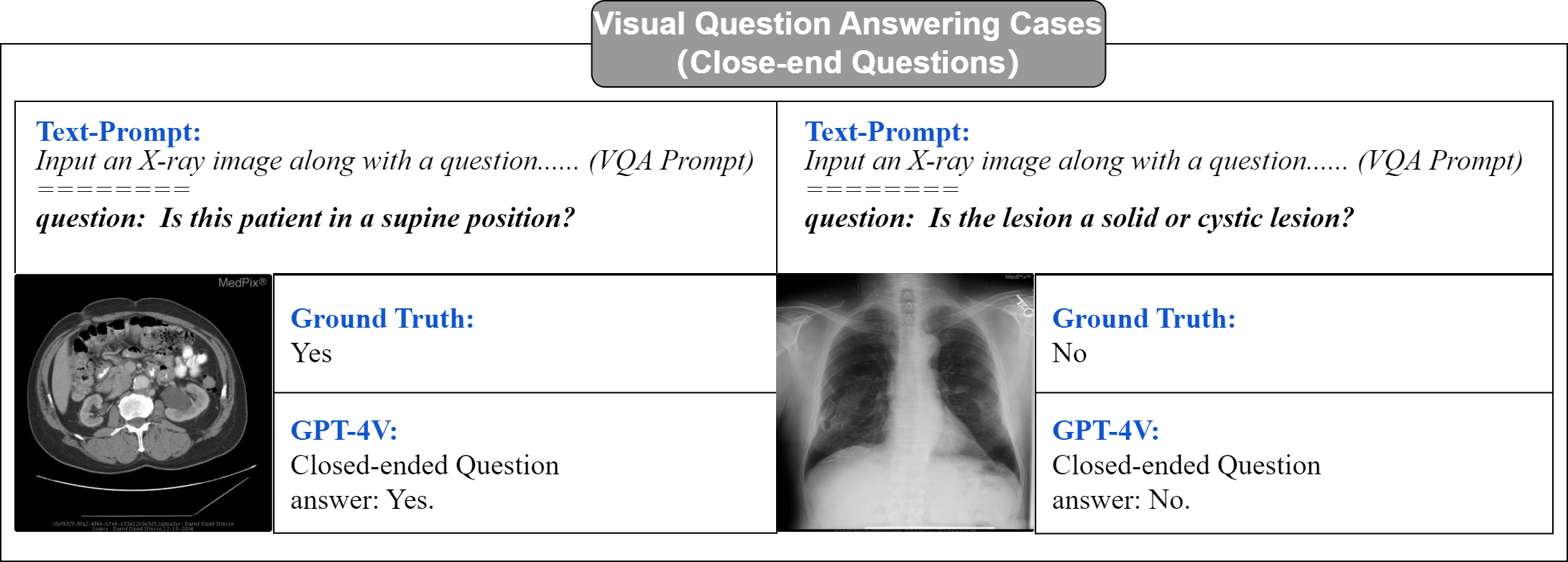}
\caption{VQA case examples for close-end questions. By few-shot prompts, GPT-4V could discern the question type and generate correct ``yes" or ``no" answers.}
\label{fig:Close-end VQA example}
\end{figure}

\begin{figure}[h]
\includegraphics[width=0.5\textwidth]{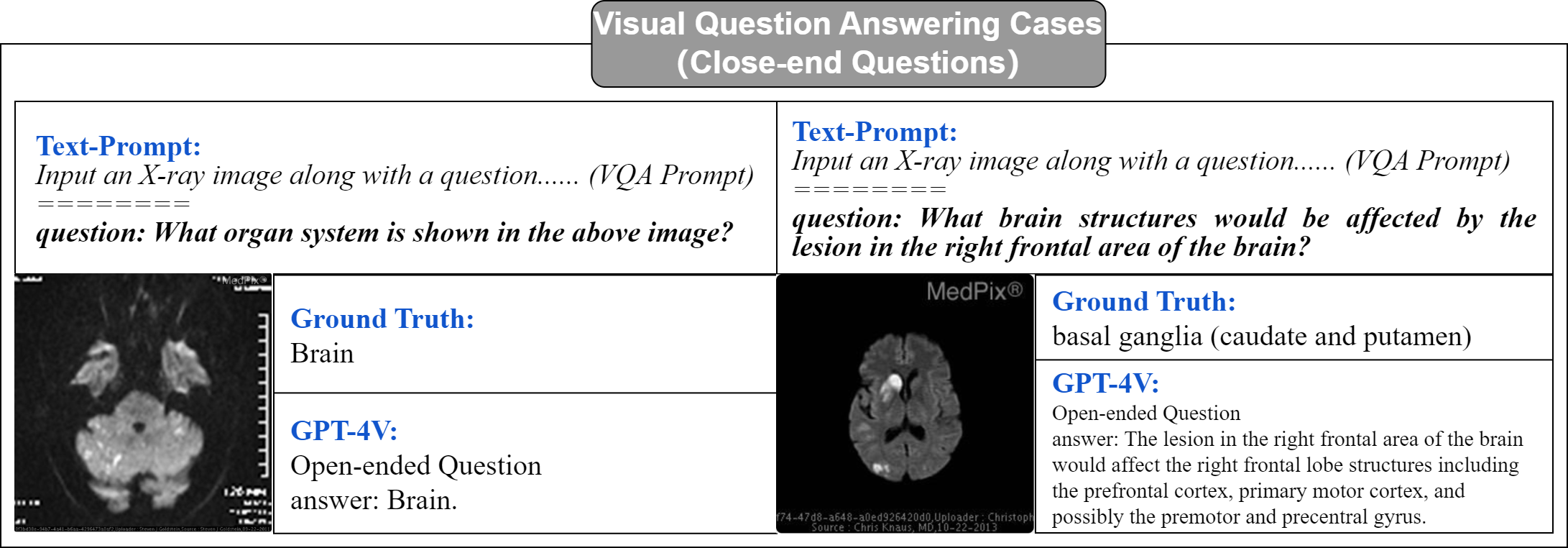}
\caption{VQA case examples for open-end questions. By few-shot prompts, GPT-4V has the capacity to generate correct answers (left). However, it may sometimes overlook visual information and generate answers solely based on general medical knowledge (right).}
\label{fig:Open-end VQA example}
\end{figure}

We delve into specific cases of VQA illustrated in Figures~\ref{fig:Close-end VQA example}, and~\ref{fig:Open-end VQA example}. Figure \ref{fig:Close-end VQA example} presents two examples of the close-end questions with a single-word answer of ``yes" or ``no". Effectively addressing such questions requires GPT-4V to accomplish two key tasks: firstly, discern that the question is close-end, and secondly, comprehend the attributes of a close-end question. This proficiency is crucial because, despite the simplicity of these questions, GPT-4V tends to generate more extended responses, which will be deemed incorrect when assessing the accuracy of answers to close-end questions. Thus, GPT-4V faces the challenge of identifying when a question demands a concise one-word answer and adjusting its response generation accordingly. Properly tailoring its output to match the question type is essential for GPT-4V's accuracy in dealing with close-end questions.

% The open-end questions are which need to generate other than just "yes" or "no" answers. So, it will have more freedom. However, for the VQA task, the answers are those which already exist in the answer pools. So, it will more hard for GPT-4V to generate exactly same answer sentence. We listed two samples in Figure \ref{fig:Open-end VQA example}, we can see the first one has correct answer, because the answer is pretty straight forward and just with one word. The second example has wrong answer, from the answer it self, the answer the GPT-4V given is "the lesion in the right frontal area of the brain would affect the right frontal lobe structres including the prefrontal cortext, primary motor cortex, and possibly the premotor and precentral gyrus", which seems just given the knowledge of "what brain structures would be affected by the lesion in the right frontal area of the brain?", because this question seems like a question only need to search the knowledge without involve the image, so the gpt-4v did this without looking in the picture.

Open-end questions demand more intricate responses beyond a simple ``yes" or ``no", providing flexibility in generating answers. However, for the VQA task, the challenge lies in predefined answer pools, making it hard for GPT-4V to generate an exact sentence that matches these predefined answers. In Figure \ref{fig:Open-end VQA example}, we present two examples of open-end questions. The first example showcases a correct, succinct response from GPT-4V, consisting of just one word. This showcases GPT-4V's effective handling of simple, direct open-end questions. The second example, however, depicts an incorrect answer by GPT-4V. The response provided by GPT-4V is a detailed explanation of the brain structures potentially affected by a lesion in the right frontal area, encompassing the prefrontal cortex, primary motor cortex, and possibly the premotor and precentral gyrus. This response appears more fitting for a theoretical question about brain anatomy, such as ``What brain structures would be affected by a lesion in the right frontal area of the brain?"
In this case, GPT-4V seems to have relied solely on medical knowledge without adequately considering the specific image in question. This suggests that GPT-4V might sometimes overlook the visual information in VQA tasks, especially when the question can be addressed through general medical knowledge. This insight underscores the challenge GPT-4V faces to generate responses for open-end VQA tasks.\\

\noindent\textbf{Visual Grounding Cases}\\

\begin{figure}[h]
\includegraphics[width=0.5\textwidth]{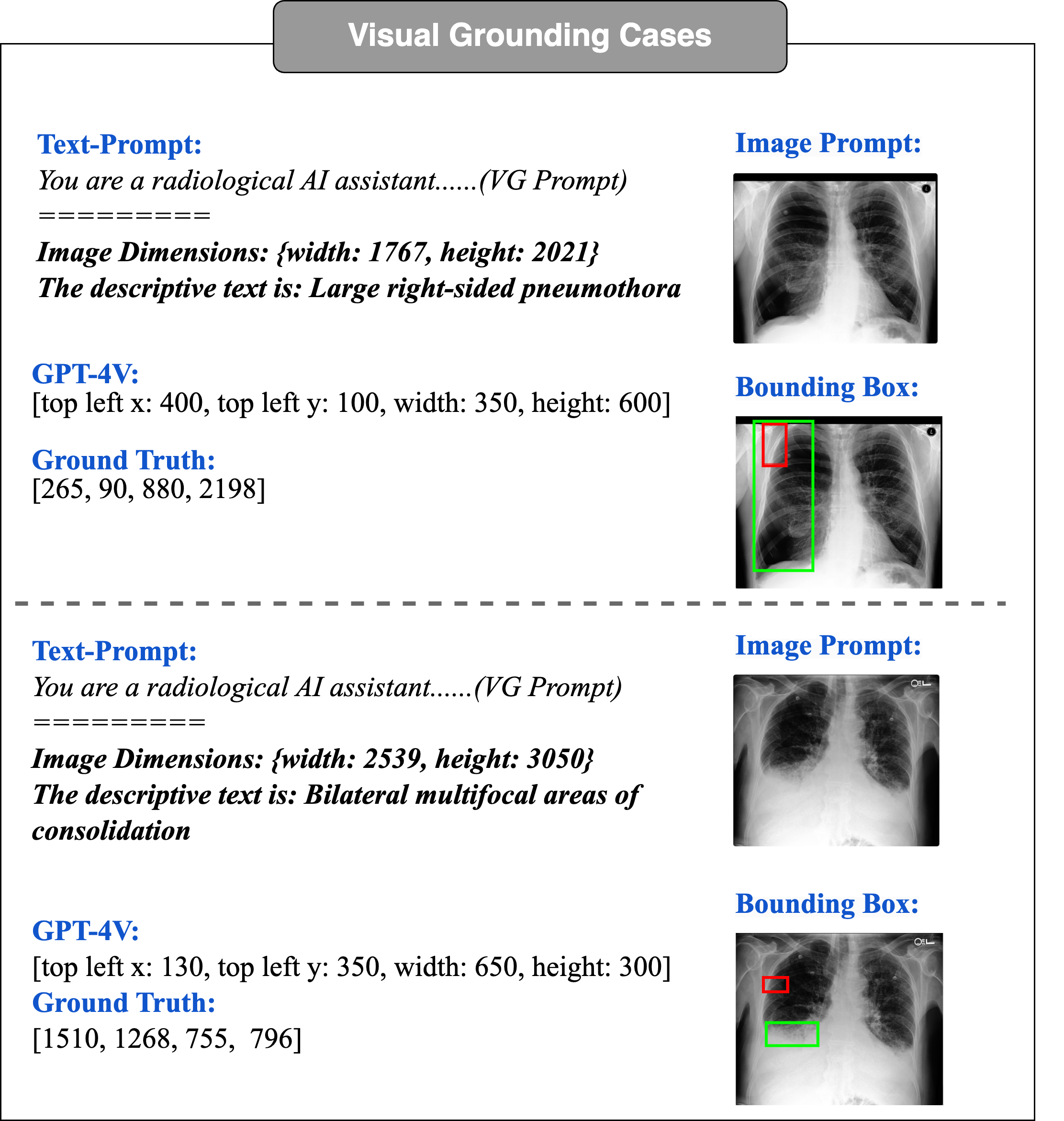}
\caption{Visual Grounding Prompt case. The bounding boxes in red are predicted by GPT-4V, while the ground truth bounding boxes are in green.}
\label{fig:Visual Grounding prompt examples}
\end{figure}

In Figure \ref{fig:Visual Grounding prompt examples} we provide two examples that shed light on GPT-4V's performance in visual grounding tasks. Our examination suggests that while GPT-4V exhibits the capability to generate bounding boxes, its performance is suboptimal, especially in precisely locating objects within images. This limitation might stem from GPT-4V's challenges in processing and interpreting detailed image information, particularly in the context of medical images where a nuanced focus on fine-grid features is crucial for accurate visual grounding. Furthermore, we posit that GPT-4V's training, predominantly on common images, may contribute to its suboptimal performance with medical images. The model's limited exposure to diverse and labelled medical data might be a key factor in this inadequacy. 

% \begin{figure}[h]
% \includegraphics[width=0.5\textwidth]{VG_Case/VG_CASE1.png}
% \caption{Visual Grounding Prompt examples. The bounding boxes in red color are predicted boxes by GPT-4V, and the green bounding boxes are ground truth boxes. GPT-4V is capable of generating and estimating the bounding box coordinates for the reference text within an image. However, the results show that the GPT-4V cannot understand medical images properly.}
% \label{fig:Visual Grounding prompt examples}
% \end{figure}

\section{Summary and discussion}

This study thoroughly investigates GPT-4V's performance in three key medical applications, employing a comprehensive evaluation methodology encompassing quantitative analysis, human assessment, and case studies.

%In the context of the radiology report generation task, GPT-4V effectively generates radiology reports from medical images, matching specialized models in language fluency and clinical relevance. Human evaluations reveal its higher accuracy and detail, suggesting conventional metrics fall short in assessing large language models like GPT-4V. 
In the domain of radiology report generation, GPT-4V effectively generates radiology reports from medical images, close to models specially trained for this task in language fluency and clinical relevance. Human evaluations reveal its higher accuracy and detailed generation, suggesting conventional metrics fall short in assessing LLMs like GPT-4V. Beyond quantitative analysis, notable observations include the impact of different prompts on GPT-4V's report generation. Specifically, a mixed-example prompt, incorporating both normal and abnormal cases, enhances report quality compared to prompts with solely normal or abnormal examples. Also, it is found that GPT-4V's ability to generate additional information not present in the ground truth, though visible in images, contributes to relatively lower BLEU scores that focus on word-matching rates, compared to specially trained R2Gen models. Remarkably, GPT-4V exhibits a notably lower CIDEr score than conventional R2Gen models, consistent with findings from R2GenGPT~\citep{wang2023r2gengpt}, which employed a frozen LLM for medical report generation. This lower CIDEr score may suggest a tendency of LLMs to generate repetitive description patterns. Meanwhile, when evaluating the rank correlation between NLP/clinical scores and the human rating, the latter aligns more closely with clinical scores than NLP scores, consistent with the observation that GPT-4V exhibits competitive clinical efficacy close to high-performing R2Gen models.

%we compare the performance of GPT-4V with SOTA methods in radiology report generation using the MIMIC-CXR dataset. The comparison includes standard image captioning models and specialized radiology report generation models. GPT-4V, despite being a general-purpose model, demonstrates impressive capabilities in generating medical reports, especially when using few-shot mixed-example prompts. Interestingly, we find that GPT-4V can generate information not present in the reference data but visible in the images, leading to lower scores in metrics like BLEU that focus on word-match rates. The clinical efficacy of GPT-4V on the MIMIC-CXR dataset is also assessed, showing competitive results with a precision of 0.353, a recall of 0.365, and an F1 score of 0.330. These results are close to those of high-performing models like METransformer and R2GenGPT.

In the domain of medical visual question answering, GPT-4V exhibits promise but achieves modest scores when compared to advanced methods, primarily due to the low correlation of its answers with standard benchmarks. It is noteworthy that human assessments consistently rate its accuracy higher than these metrics suggest. Meanwhile, several limitations associated with the VQA datasets and evaluation call for attention.
The current public datasets for VQA are constrained by a fixed answer format, imposing limitations on the training potential of models like GPT-4V and introducing challenges in accurately evaluating their performance. The inherent rigidity of these datasets becomes apparent when assessing generative models for VQA, where a more flexible evaluation approach to accommodate a spectrum of correct answers is warranted. 
%Current datasets often feature a predefined set of correct answers, creating a disparity wherein a semantically correct answer may be unfairly marked as incorrect if it doesn't precisely match the designated answers' wording.
%However, several limitations associated with the VQA dataset and model base need consideration. The current public datasets used for VQA have a fixed answer format. This rigid structure not only limits the training potential of models like GPT-4V but also presents challenges in accurately evaluating their performance. In the realm of VQA, where generative models are more appropriate, the evaluation should ideally accommodate a range of correct answers. However, the datasets often contain only a few pre-determined correct answers. This poses a problem: if a model generates an answer that is semantically correct but doesn't match the exact wording of the designated answers in the dataset, it is unfairly marked as incorrect by current evaluation methods. This rigid approach overlooks the nuanced understanding that generative models bring to VQA tasks.
%
Moreover, the predominant method for evaluating VQA tasks, rooted in classification techniques, centers around accuracy. This approach becomes less effective when transitioning to large models like LLMs for VQA tasks. Compared with traditional methods, LLMs generate answers in a more flexible and nuanced manner, leading to a variety of correct answers that may not align with the limited set of predefined ground truth and thus could be unfairly marked as incorrect by conventional evaluation metrics.
%the predominant method for evaluating VQA tasks is currently based on accuracy. This is largely because many existing approaches, which do not utilise Large Language Models (LLMs), are grounded in classification techniques. In such cases, determining the accuracy of an answer is straightforward – it either matches the pre-defined correct answer or it doesn't. However, when we shift to using large models like LLMs for VQA tasks, the evaluation criteria need to be refined. LLMs operate differently from traditional classification-based methods. They generate answers in a more flexible and nuanced way, often producing a variety of correct answers that might not exactly match the limited set of predefined correct answers in the dataset.
Consequently, there is a pressing need for more sophisticated evaluation methods, which should be adept at assessing contextual and semantic correctness, departing from the conventional practice of exact matches with predefined correct answers. Such an evolved approach would provide a more accurate gauge of LLMs' capabilities in comprehending and responding to the intricacies of VQA tasks, thereby reflecting the true potential of these advanced models in handling complex question-answering scenarios.

%Therefore, to effectively evaluate the performance of LLMs in VQA, we need to adopt more sophisticated evaluation methods. These methods should be capable of assessing whether an answer is contextually and semantically correct, rather than just checking for an exact match with a limited set of correct answers. Such an approach would provide a more accurate measure of the LLMs' capabilities in understanding and responding to VQA tasks, reflecting the true potential of these advanced models in handling complex question-answering scenarios.

In the domain of medical visual grounding, GPT-4V exhibits a need for refinement in accurately locating and identifying specific image elements, echoing a recognized gap observed in generic image tasks. This underscores a notable area for potential improvement.
%GPT-4V needs improvement in locating and identifying specific image elements, a gap also noted in generic image tasks. This highlights a potential area for enhancement. 
A recent publication ~\cite{yang2023set} introduced the Set-of-Mark prompting method, offering a novel approach to enhance the visual grounding capabilities of large multimodal models, including GPT-4V. The SoM method entails partitioning an image into regions and overlaying them with marks, such as alphanumeric characters, masks, or boxes. These marks serve to augment GPT-4V's understanding and responsiveness to visual content, leading to a significant enhancement in performance, particularly in fine-grained vision tasks within a zero-shot setting. This innovative method holds promise for addressing the identified need for improvement in GPT-4V's performance in medical image visual grounding. In the medical context, where precise identification and description of specific regions or features in images are crucial, applying the SoM technique could potentially empower models like GPT-4V to provide more accurate and detailed descriptions of medical images, thereby contributing to improved diagnosis and treatment planning.
%This new visual prompting approach enhances the visual grounding capabilities of large multimodal models like GPT-4V. It involves partitioning an image into regions and overlaying them with marks such as alphanumeric characters, masks, or boxes. These marks help GPT-4V better understand and respond to visual content, significantly improving performance on fine-grained vision tasks in a zero-shot setting. This method could be especially useful in medical image visual grounding, where precise identification and description of specific regions or features in medical images are crucial. By applying SoM, a model like GPT-4V could potentially provide more accurate and detailed descriptions of medical images, aiding in diagnosis and treatment planning.

Overall, our evaluation, incorporating standard metrics and medical expert reviews, underscores the limitations of traditional methods in fully capturing GPT-4V's capabilities, emphasizing the imperative for improved evaluation techniques tailored to large language models.

% The findings reveal that GPT-4V excels in text generation for multimodal applications beyond current models. However, traditional evaluation methods alone do not fully showcase GPT-4V's advantages, indicating the necessity for more comprehensive assessment methods. For the VQA task, The current datasets have certain limitations in evaluating results. Besides some issues where answers are provided based solely on the question, without considering image information, this might be improved by modifying the prompt. Other than these, the rest of the answers appear to be quite reasonable. Additionally, GPT-4V's ability to accurately locate objects in images is identified as an area needing improvement, which presents a direction for future enhancement. 

\section{Conclusion and limitation}
% \noindent This study provides a comprehensive evaluation of the GPT-4V model within the medical field, encompassing three primary applications. Our analysis demonstrates that GPT-4V significantly advances text generation capabilities in multi-modal contexts, surpassing current models. The model's performance, particularly in generating nuanced and contextually relevant text, highlights its potential in revolutionizing medical applications that require a deep understanding of both text and imagery. However, our findings also underscore a crucial aspect: traditional evaluation methods are insufficient to capture the full extent of GPT-4V's capabilities. This revelation points to the need for developing more robust and encompassing assessment techniques that can better evaluate the complexities of AI models in multi-modal tasks. 
% 1. sample example number
% 2. human rate 3 stage

Our comprehensive evaluation of GPT-4V across key medical applications sheds light on its remarkable capabilities and areas for refinement. GPT-4V substantially enhances text generation capabilities in multi-modal contexts. Its proficiency in generating nuanced and contextually relevant text underscores its potential to revolutionize medical applications that demand a profound understanding of both text and imagery.

While our study provides valuable insights, certain limitations present opportunities for enhancement in future research. First, our human evaluation, based on 100 cases, aligns with similar studies in the literature; however, the sample size remains relatively small, offering a representative but limited perspective on the issues discussed. %While a larger sample could enhance the robustness of our findings, constraints in manpower restricted us to this subset.
Second, our human evaluation employs a grading system with three basic levels. This categorization, while providing a general understanding, is somewhat rudimentary. Future studies could benefit from a more detailed and nuanced human evaluation framework, potentially offering more granular insights into model performance and efficiency in the assessment process. Further refinement in evaluation methodologies would contribute to a more comprehensive understanding of the intricacies involved in model assessments.

\bibliographystyle{model2-names.bst}\biboptions{authoryear}
\bibliography{refs}

\begin{thebibliography}{67}
\expandafter\ifx\csname natexlab\endcsname\relax\def\natexlab#1{#1}\fi
\providecommand{\url}[1]{\texttt{#1}}
\providecommand{\href}[2]{#2}
\providecommand{\path}[1]{#1}
\providecommand{\DOIprefix}{doi:}
\providecommand{\ArXivprefix}{arXiv:}
\providecommand{\URLprefix}{URL: }
\providecommand{\Pubmedprefix}{pmid:}
\providecommand{\doi}[1]{\href{http://dx.doi.org/#1}{\path{#1}}}
\providecommand{\Pubmed}[1]{\href{pmid:#1}{\path{#1}}}
\providecommand{\bibinfo}[2]{#2}
\ifx\xfnm\relax \def\xfnm[#1]{\unskip,\space#1}\fi
%Type = Inproceedings
\bibitem[{Ambati and Dudyala(2018)}]{ambati2018sequence}
\bibinfo{author}{Ambati, R.}, \bibinfo{author}{Dudyala, C.R.}, \bibinfo{year}{2018}.
\newblock \bibinfo{title}{A sequence-to-sequence model approach for imageclef 2018 medical domain visual question answering}, in: \bibinfo{booktitle}{2018 15th IEEE India Council International Conference (INDICON)}, \bibinfo{organization}{IEEE}. pp. \bibinfo{pages}{1--6}.
%Type = Article
\bibitem[{Anil et~al.(2023)Anil, Dai, Firat, Johnson, Lepikhin, Passos, Shakeri, Taropa, Bailey, Chen et~al.}]{anil2023palm}
\bibinfo{author}{Anil, R.}, \bibinfo{author}{Dai, A.M.}, \bibinfo{author}{Firat, O.}, \bibinfo{author}{Johnson, M.}, \bibinfo{author}{Lepikhin, D.}, \bibinfo{author}{Passos, A.}, \bibinfo{author}{Shakeri, S.}, \bibinfo{author}{Taropa, E.}, \bibinfo{author}{Bailey, P.}, \bibinfo{author}{Chen, Z.}, et~al., \bibinfo{year}{2023}.
\newblock \bibinfo{title}{Palm 2 technical report}.
\newblock \bibinfo{journal}{arXiv preprint arXiv:2305.10403} .
%Type = Article
\bibitem[{Awadalla et~al.(2023)Awadalla, Gao, Gardner, Hessel, Hanafy, Zhu, Marathe, Bitton, Gadre, Sagawa et~al.}]{awadalla2023openflamingo}
\bibinfo{author}{Awadalla, A.}, \bibinfo{author}{Gao, I.}, \bibinfo{author}{Gardner, J.}, \bibinfo{author}{Hessel, J.}, \bibinfo{author}{Hanafy, Y.}, \bibinfo{author}{Zhu, W.}, \bibinfo{author}{Marathe, K.}, \bibinfo{author}{Bitton, Y.}, \bibinfo{author}{Gadre, S.}, \bibinfo{author}{Sagawa, S.}, et~al., \bibinfo{year}{2023}.
\newblock \bibinfo{title}{Openflamingo: An open-source framework for training large autoregressive vision-language models}.
\newblock \bibinfo{journal}{arXiv preprint arXiv:2308.01390} .
%Type = Inproceedings
\bibitem[{Banerjee and Lavie(2005)}]{banerjee-lavie-2005-meteor}
\bibinfo{author}{Banerjee, S.}, \bibinfo{author}{Lavie, A.}, \bibinfo{year}{2005}.
\newblock \bibinfo{title}{{METEOR:} an automatic metric for {MT} evaluation with improved correlation with human judgments}, in: \bibinfo{editor}{Goldstein, J.}, \bibinfo{editor}{Lavie, A.}, \bibinfo{editor}{Lin, C.}, \bibinfo{editor}{Voss, C.R.} (Eds.), \bibinfo{booktitle}{Proceedings of the Workshop on Intrinsic and Extrinsic Evaluation Measures for Machine Translation and/or Summarization@ACL 2005, Ann Arbor, Michigan, USA, June 29, 2005}, \bibinfo{publisher}{Association for Computational Linguistics}. pp. \bibinfo{pages}{65--72}.
\newblock \URLprefix \url{https://aclanthology.org/W05-0909/}.
%Type = Inproceedings
\bibitem[{Bannur et~al.(2023)Bannur, Hyland, Liu, Perez-Garcia, Ilse, Castro, Boecking, Sharma, Bouzid, Thieme et~al.}]{bannur2023learning}
\bibinfo{author}{Bannur, S.}, \bibinfo{author}{Hyland, S.}, \bibinfo{author}{Liu, Q.}, \bibinfo{author}{Perez-Garcia, F.}, \bibinfo{author}{Ilse, M.}, \bibinfo{author}{Castro, D.C.}, \bibinfo{author}{Boecking, B.}, \bibinfo{author}{Sharma, H.}, \bibinfo{author}{Bouzid, K.}, \bibinfo{author}{Thieme, A.}, et~al., \bibinfo{year}{2023}.
\newblock \bibinfo{title}{Learning to exploit temporal structure for biomedical vision-language processing}, in: \bibinfo{booktitle}{Proceedings of the IEEE/CVF Conference on Computer Vision and Pattern Recognition}, pp. \bibinfo{pages}{15016--15027}.
%Type = Inproceedings
\bibitem[{Boecking et~al.(2022)Boecking, Usuyama, Bannur, Castro, Schwaighofer, Hyland, Wetscherek, Naumann, Nori, Alvarez-Valle et~al.}]{boecking2022making}
\bibinfo{author}{Boecking, B.}, \bibinfo{author}{Usuyama, N.}, \bibinfo{author}{Bannur, S.}, \bibinfo{author}{Castro, D.C.}, \bibinfo{author}{Schwaighofer, A.}, \bibinfo{author}{Hyland, S.}, \bibinfo{author}{Wetscherek, M.}, \bibinfo{author}{Naumann, T.}, \bibinfo{author}{Nori, A.}, \bibinfo{author}{Alvarez-Valle, J.}, et~al., \bibinfo{year}{2022}.
\newblock \bibinfo{title}{Making the most of text semantics to improve biomedical vision--language processing}, in: \bibinfo{booktitle}{European conference on computer vision}, \bibinfo{organization}{Springer}. pp. \bibinfo{pages}{1--21}.
%Type = Inproceedings
\bibitem[{Chen et~al.(2021)Chen, Shen, Song and Wan}]{chen2022cross}
\bibinfo{author}{Chen, Z.}, \bibinfo{author}{Shen, Y.}, \bibinfo{author}{Song, Y.}, \bibinfo{author}{Wan, X.}, \bibinfo{year}{2021}.
\newblock \bibinfo{title}{Cross-modal memory networks for radiology report generation}, in: \bibinfo{editor}{Zong, C.}, \bibinfo{editor}{Xia, F.}, \bibinfo{editor}{Li, W.}, \bibinfo{editor}{Navigli, R.} (Eds.), \bibinfo{booktitle}{Proceedings of the 59th Annual Meeting of the Association for Computational Linguistics and the 11th International Joint Conference on Natural Language Processing, {ACL/IJCNLP} 2021, (Volume 1: Long Papers), Virtual Event, August 1-6, 2021}, \bibinfo{publisher}{Association for Computational Linguistics}. pp. \bibinfo{pages}{5904--5914}.
\newblock \URLprefix \url{https://doi.org/10.18653/v1/2021.acl-long.459}, \DOIprefix\doi{10.18653/v1/2021.acl-long.459}.
%Type = Inproceedings
\bibitem[{Chen et~al.(2020)Chen, Song, Chang and Wan}]{chen2020generating}
\bibinfo{author}{Chen, Z.}, \bibinfo{author}{Song, Y.}, \bibinfo{author}{Chang, T.}, \bibinfo{author}{Wan, X.}, \bibinfo{year}{2020}.
\newblock \bibinfo{title}{Generating radiology reports via memory-driven transformer}, in: \bibinfo{editor}{Webber, B.}, \bibinfo{editor}{Cohn, T.}, \bibinfo{editor}{He, Y.}, \bibinfo{editor}{Liu, Y.} (Eds.), \bibinfo{booktitle}{Proceedings of the 2020 Conference on Empirical Methods in Natural Language Processing, {EMNLP} 2020, Online, November 16-20, 2020}, \bibinfo{publisher}{Association for Computational Linguistics}. pp. \bibinfo{pages}{1439--1449}.
\newblock \URLprefix \url{https://doi.org/10.18653/v1/2020.emnlp-main.112}, \DOIprefix\doi{10.18653/v1/2020.emnlp-main.112}.
%Type = Inproceedings
\bibitem[{Chen et~al.(2023)Chen, Zhou, Tran, Zhao, Wan, Ooi, Cheng, Thng, Xu, Liu et~al.}]{chen2023medical}
\bibinfo{author}{Chen, Z.}, \bibinfo{author}{Zhou, Y.}, \bibinfo{author}{Tran, A.}, \bibinfo{author}{Zhao, J.}, \bibinfo{author}{Wan, L.}, \bibinfo{author}{Ooi, G.S.K.}, \bibinfo{author}{Cheng, L.T.E.}, \bibinfo{author}{Thng, C.H.}, \bibinfo{author}{Xu, X.}, \bibinfo{author}{Liu, Y.}, et~al., \bibinfo{year}{2023}.
\newblock \bibinfo{title}{Medical phrase grounding with region-phrase context contrastive alignment}, in: \bibinfo{booktitle}{International Conference on Medical Image Computing and Computer-Assisted Intervention}, \bibinfo{organization}{Springer}. pp. \bibinfo{pages}{371--381}.
%Type = Inproceedings
\bibitem[{Cornia et~al.(2020)Cornia, Stefanini, Baraldi and Cucchiara}]{cornia2020meshedmemory}
\bibinfo{author}{Cornia, M.}, \bibinfo{author}{Stefanini, M.}, \bibinfo{author}{Baraldi, L.}, \bibinfo{author}{Cucchiara, R.}, \bibinfo{year}{2020}.
\newblock \bibinfo{title}{Meshed-memory transformer for image captioning}, in: \bibinfo{booktitle}{2020 {IEEE/CVF} Conference on Computer Vision and Pattern Recognition, {CVPR} 2020, Seattle, WA, USA, June 13-19, 2020}, \bibinfo{publisher}{Computer Vision Foundation / {IEEE}}. pp. \bibinfo{pages}{10575--10584}.
\newblock \URLprefix \url{https://openaccess.thecvf.com/content\_CVPR\_2020/html/Cornia\_Meshed-Memory\_Transformer\_for\_Image\_Captioning\_CVPR\_2020\_paper.html}, \DOIprefix\doi{10.1109/CVPR42600.2020.01059}.
%Type = Article
\bibitem[{Dai et~al.(2022)Dai, Sun, Dong, Hao, Sui and Wei}]{dai2022can}
\bibinfo{author}{Dai, D.}, \bibinfo{author}{Sun, Y.}, \bibinfo{author}{Dong, L.}, \bibinfo{author}{Hao, Y.}, \bibinfo{author}{Sui, Z.}, \bibinfo{author}{Wei, F.}, \bibinfo{year}{2022}.
\newblock \bibinfo{title}{Why can gpt learn in-context? language models secretly perform gradient descent as meta optimizers}.
\newblock \bibinfo{journal}{arXiv preprint arXiv:2212.10559} .
%Type = Inproceedings
\bibitem[{Deng et~al.(2021)Deng, Yang, Chen, Zhou and Li}]{deng2021transvg}
\bibinfo{author}{Deng, J.}, \bibinfo{author}{Yang, Z.}, \bibinfo{author}{Chen, T.}, \bibinfo{author}{Zhou, W.}, \bibinfo{author}{Li, H.}, \bibinfo{year}{2021}.
\newblock \bibinfo{title}{Transvg: End-to-end visual grounding with transformers}, in: \bibinfo{booktitle}{Proceedings of the IEEE/CVF International Conference on Computer Vision}, pp. \bibinfo{pages}{1769--1779}.
%Type = Inproceedings
\bibitem[{Do et~al.(2021)Do, Nguyen, Tjiputra, Tran, Tran and Nguyen}]{do2021multiple}
\bibinfo{author}{Do, T.}, \bibinfo{author}{Nguyen, B.X.}, \bibinfo{author}{Tjiputra, E.}, \bibinfo{author}{Tran, M.}, \bibinfo{author}{Tran, Q.D.}, \bibinfo{author}{Nguyen, A.}, \bibinfo{year}{2021}.
\newblock \bibinfo{title}{Multiple meta-model quantifying for medical visual question answering}, in: \bibinfo{booktitle}{International Conference on Medical Image Computing and Computer-Assisted Intervention}, \bibinfo{organization}{Springer}. pp. \bibinfo{pages}{64--74}.
%Type = Inproceedings
\bibitem[{Du et~al.(2022)Du, Fu, Liu and Wang}]{du2022visual}
\bibinfo{author}{Du, Y.}, \bibinfo{author}{Fu, Z.}, \bibinfo{author}{Liu, Q.}, \bibinfo{author}{Wang, Y.}, \bibinfo{year}{2022}.
\newblock \bibinfo{title}{Visual grounding with transformers}, in: \bibinfo{booktitle}{2022 IEEE International Conference on Multimedia and Expo (ICME)}, \bibinfo{organization}{IEEE}. pp. \bibinfo{pages}{1--6}.
%Type = Article
\bibitem[{Eslami et~al.(2021)Eslami, de~Melo and Meinel}]{eslami2021does}
\bibinfo{author}{Eslami, S.}, \bibinfo{author}{de~Melo, G.}, \bibinfo{author}{Meinel, C.}, \bibinfo{year}{2021}.
\newblock \bibinfo{title}{Does clip benefit visual question answering in the medical domain as much as it does in the general domain?}
\newblock \bibinfo{journal}{arXiv preprint arXiv:2112.13906} .
%Type = Inproceedings
\bibitem[{Finn et~al.(2017)Finn, Abbeel and Levine}]{finn2017model}
\bibinfo{author}{Finn, C.}, \bibinfo{author}{Abbeel, P.}, \bibinfo{author}{Levine, S.}, \bibinfo{year}{2017}.
\newblock \bibinfo{title}{Model-agnostic meta-learning for fast adaptation of deep networks}, in: \bibinfo{booktitle}{International conference on machine learning}, \bibinfo{organization}{PMLR}. pp. \bibinfo{pages}{1126--1135}.
%Type = Article
\bibitem[{He et~al.(2020)He, Zhang, Mou, Xing and Xie}]{he2020pathvqa}
\bibinfo{author}{He, X.}, \bibinfo{author}{Zhang, Y.}, \bibinfo{author}{Mou, L.}, \bibinfo{author}{Xing, E.}, \bibinfo{author}{Xie, P.}, \bibinfo{year}{2020}.
\newblock \bibinfo{title}{Pathvqa: 30000+ questions for medical visual question answering}.
\newblock \bibinfo{journal}{arXiv preprint arXiv:2003.10286} .
%Type = Article
\bibitem[{Huang et~al.(2023a)Huang, Zhou, Li, Yang, Liu and Wang}]{huang2023enhancing}
\bibinfo{author}{Huang, W.}, \bibinfo{author}{Zhou, H.}, \bibinfo{author}{Li, C.}, \bibinfo{author}{Yang, H.}, \bibinfo{author}{Liu, J.}, \bibinfo{author}{Wang, S.}, \bibinfo{year}{2023}a.
\newblock \bibinfo{title}{Enhancing representation in radiography-reports foundation model: A granular alignment algorithm using masked contrastive learning}.
\newblock \bibinfo{journal}{arXiv preprint arXiv:2309.05904} .
%Type = Inproceedings
\bibitem[{Huang et~al.(2023b)Huang, Zhang and Zhang}]{huang2023kiut}
\bibinfo{author}{Huang, Z.}, \bibinfo{author}{Zhang, X.}, \bibinfo{author}{Zhang, S.}, \bibinfo{year}{2023}b.
\newblock \bibinfo{title}{Kiut: Knowledge-injected u-transformer for radiology report generation}, in: \bibinfo{booktitle}{Proceedings of the IEEE/CVF Conference on Computer Vision and Pattern Recognition}, pp. \bibinfo{pages}{19809--19818}.
%Type = Inproceedings
\bibitem[{Jiang et~al.(2020)Jiang, Misra, Rohrbach, Learned-Miller and Chen}]{jiang2020defense}
\bibinfo{author}{Jiang, H.}, \bibinfo{author}{Misra, I.}, \bibinfo{author}{Rohrbach, M.}, \bibinfo{author}{Learned-Miller, E.}, \bibinfo{author}{Chen, X.}, \bibinfo{year}{2020}.
\newblock \bibinfo{title}{In defense of grid features for visual question answering}, in: \bibinfo{booktitle}{Proceedings of the IEEE/CVF Conference on Computer Vision and Pattern Recognition}, pp. \bibinfo{pages}{10267--10276}.
%Type = Article
\bibitem[{Johnson et~al.(2019)Johnson, Pollard, Berkowitz, Greenbaum, Lungren, Deng, Mark and Horng}]{2019MIMIC}
\bibinfo{author}{Johnson, A.E.W.}, \bibinfo{author}{Pollard, T.J.}, \bibinfo{author}{Berkowitz, S.J.}, \bibinfo{author}{Greenbaum, N.R.}, \bibinfo{author}{Lungren, M.P.}, \bibinfo{author}{Deng, C.}, \bibinfo{author}{Mark, R.G.}, \bibinfo{author}{Horng, S.}, \bibinfo{year}{2019}.
\newblock \bibinfo{title}{{MIMIC-CXR:} {A} large publicly available database of labeled chest radiographs}.
\newblock \bibinfo{journal}{CoRR} \bibinfo{volume}{abs/1901.07042}.
\newblock \URLprefix \url{http://arxiv.org/abs/1901.07042}, \href{http://arxiv.org/abs/1901.07042}{\tt arXiv:1901.07042}.
%Type = Inproceedings
\bibitem[{Kamath et~al.(2021)Kamath, Singh, LeCun, Synnaeve, Misra and Carion}]{kamath2021mdetr}
\bibinfo{author}{Kamath, A.}, \bibinfo{author}{Singh, M.}, \bibinfo{author}{LeCun, Y.}, \bibinfo{author}{Synnaeve, G.}, \bibinfo{author}{Misra, I.}, \bibinfo{author}{Carion, N.}, \bibinfo{year}{2021}.
\newblock \bibinfo{title}{Mdetr-modulated detection for end-to-end multi-modal understanding}, in: \bibinfo{booktitle}{Proceedings of the IEEE/CVF International Conference on Computer Vision}, pp. \bibinfo{pages}{1780--1790}.
%Type = Inproceedings
\bibitem[{Khare et~al.(2021)Khare, Bagal, Mathew, Devi, Priyakumar and Jawahar}]{khare2021mmbert}
\bibinfo{author}{Khare, Y.}, \bibinfo{author}{Bagal, V.}, \bibinfo{author}{Mathew, M.}, \bibinfo{author}{Devi, A.}, \bibinfo{author}{Priyakumar, U.D.}, \bibinfo{author}{Jawahar, C.}, \bibinfo{year}{2021}.
\newblock \bibinfo{title}{Mmbert: multimodal bert pretraining for improved medical vqa}, in: \bibinfo{booktitle}{2021 IEEE 18th International Symposium on Biomedical Imaging (ISBI)}, \bibinfo{organization}{IEEE}. pp. \bibinfo{pages}{1033--1036}.
%Type = Article
\bibitem[{Lau et~al.(2018)Lau, Gayen, Ben~Abacha and Demner-Fushman}]{lau2018dataset}
\bibinfo{author}{Lau, J.J.}, \bibinfo{author}{Gayen, S.}, \bibinfo{author}{Ben~Abacha, A.}, \bibinfo{author}{Demner-Fushman, D.}, \bibinfo{year}{2018}.
\newblock \bibinfo{title}{A dataset of clinically generated visual questions and answers about radiology images}.
\newblock \bibinfo{journal}{Scientific data} \bibinfo{volume}{5}, \bibinfo{pages}{1--10}.
%Type = Article
\bibitem[{Li et~al.(2023a)Li, Wong, Zhang, Usuyama, Liu, Yang, Naumann, Poon and Gao}]{li2023llava}
\bibinfo{author}{Li, C.}, \bibinfo{author}{Wong, C.}, \bibinfo{author}{Zhang, S.}, \bibinfo{author}{Usuyama, N.}, \bibinfo{author}{Liu, H.}, \bibinfo{author}{Yang, J.}, \bibinfo{author}{Naumann, T.}, \bibinfo{author}{Poon, H.}, \bibinfo{author}{Gao, J.}, \bibinfo{year}{2023}a.
\newblock \bibinfo{title}{Llava-med: Training a large language-and-vision assistant for biomedicine in one day}.
\newblock \bibinfo{journal}{arXiv preprint arXiv:2306.00890} .
%Type = Article
\bibitem[{Li et~al.(2023b)Li, Li, Savarese and Hoi}]{li2023blip}
\bibinfo{author}{Li, J.}, \bibinfo{author}{Li, D.}, \bibinfo{author}{Savarese, S.}, \bibinfo{author}{Hoi, S.}, \bibinfo{year}{2023}b.
\newblock \bibinfo{title}{Blip-2: Bootstrapping language-image pre-training with frozen image encoders and large language models}.
\newblock \bibinfo{journal}{arXiv preprint arXiv:2301.12597} .
%Type = Inproceedings
\bibitem[{Li et~al.(2023c)Li, Lin, Chen, Lin, Liang and Chang}]{li2023dynamic}
\bibinfo{author}{Li, M.}, \bibinfo{author}{Lin, B.}, \bibinfo{author}{Chen, Z.}, \bibinfo{author}{Lin, H.}, \bibinfo{author}{Liang, X.}, \bibinfo{author}{Chang, X.}, \bibinfo{year}{2023}c.
\newblock \bibinfo{title}{Dynamic graph enhanced contrastive learning for chest x-ray report generation}, in: \bibinfo{booktitle}{Proceedings of the IEEE/CVF Conference on Computer Vision and Pattern Recognition}, pp. \bibinfo{pages}{3334--3343}.
%Type = Article
\bibitem[{Li and Sigal(2021)}]{li2021referring}
\bibinfo{author}{Li, M.}, \bibinfo{author}{Sigal, L.}, \bibinfo{year}{2021}.
\newblock \bibinfo{title}{Referring transformer: A one-step approach to multi-task visual grounding}.
\newblock \bibinfo{journal}{Advances in neural information processing systems} \bibinfo{volume}{34}, \bibinfo{pages}{19652--19664}.
%Type = Inproceedings
\bibitem[{Li et~al.(2018)Li, Liang, Hu and Xing}]{li2018hybrid}
\bibinfo{author}{Li, Y.}, \bibinfo{author}{Liang, X.}, \bibinfo{author}{Hu, Z.}, \bibinfo{author}{Xing, E.P.}, \bibinfo{year}{2018}.
\newblock \bibinfo{title}{Hybrid retrieval-generation reinforced agent for medical image report generation}, in: \bibinfo{editor}{Bengio, S.}, \bibinfo{editor}{Wallach, H.M.}, \bibinfo{editor}{Larochelle, H.}, \bibinfo{editor}{Grauman, K.}, \bibinfo{editor}{Cesa{-}Bianchi, N.}, \bibinfo{editor}{Garnett, R.} (Eds.), \bibinfo{booktitle}{Advances in Neural Information Processing Systems 31: Annual Conference on Neural Information Processing Systems 2018, NeurIPS 2018, December 3-8, 2018, Montr{\'{e}}al, Canada}, pp. \bibinfo{pages}{1537--1547}.
\newblock \URLprefix \url{https://proceedings.neurips.cc/paper/2018/hash/e07413354875be01a996dc560274708e-Abstract.html}.
%Type = Article
\bibitem[{Li et~al.(2023d)Li, Wang, Hu, Chen, Zhong, Lyu and Zhang}]{li2023comprehensive}
\bibinfo{author}{Li, Y.}, \bibinfo{author}{Wang, L.}, \bibinfo{author}{Hu, B.}, \bibinfo{author}{Chen, X.}, \bibinfo{author}{Zhong, W.}, \bibinfo{author}{Lyu, C.}, \bibinfo{author}{Zhang, M.}, \bibinfo{year}{2023}d.
\newblock \bibinfo{title}{A comprehensive evaluation of gpt-4v on knowledge-intensive visual question answering}.
\newblock \bibinfo{journal}{arXiv preprint arXiv:2311.07536} .
%Type = Inproceedings
\bibitem[{Lin(2004)}]{lin-2004-rouge}
\bibinfo{author}{Lin, C.Y.}, \bibinfo{year}{2004}.
\newblock \bibinfo{title}{{ROUGE}: A package for automatic evaluation of summaries}, in: \bibinfo{booktitle}{Text Summarization Branches Out}, \bibinfo{publisher}{Association for Computational Linguistics}, \bibinfo{address}{Barcelona, Spain}. pp. \bibinfo{pages}{74--81}.
\newblock \URLprefix \url{https://aclanthology.org/W04-1013}.
%Type = Inproceedings
\bibitem[{Liu et~al.(2021a)Liu, Wu, Ge, Fan and Zou}]{liu2021exploring}
\bibinfo{author}{Liu, F.}, \bibinfo{author}{Wu, X.}, \bibinfo{author}{Ge, S.}, \bibinfo{author}{Fan, W.}, \bibinfo{author}{Zou, Y.}, \bibinfo{year}{2021}a.
\newblock \bibinfo{title}{Exploring and distilling posterior and prior knowledge for radiology report generation}, in: \bibinfo{booktitle}{Proceedings of the IEEE/CVF conference on computer vision and pattern recognition}, pp. \bibinfo{pages}{13753--13762}.
%Type = Inproceedings
\bibitem[{Liu et~al.(2021b)Liu, Wu, Ge, Fan and Zou}]{CVPR201_PPKD}
\bibinfo{author}{Liu, F.}, \bibinfo{author}{Wu, X.}, \bibinfo{author}{Ge, S.}, \bibinfo{author}{Fan, W.}, \bibinfo{author}{Zou, Y.}, \bibinfo{year}{2021}b.
\newblock \bibinfo{title}{Exploring and distilling posterior and prior knowledge for radiology report generation}, in: \bibinfo{booktitle}{{IEEE} Conference on Computer Vision and Pattern Recognition, {CVPR} 2021, virtual, June 19-25, 2021}, \bibinfo{publisher}{Computer Vision Foundation / {IEEE}}. pp. \bibinfo{pages}{13753--13762}.
\newblock \URLprefix \url{https://openaccess.thecvf.com/content/CVPR2021/html/Liu\_Exploring\_and\_Distilling\_Posterior\_and\_Prior\_Knowledge\_for\_Radiology\_Report\_CVPR\_2021\_paper.html}, \DOIprefix\doi{10.1109/CVPR46437.2021.01354}.
%Type = Inproceedings
\bibitem[{Liu et~al.(2023)Liu, Wang, Xu and Zhou}]{liu2023q2atransformer}
\bibinfo{author}{Liu, Y.}, \bibinfo{author}{Wang, Z.}, \bibinfo{author}{Xu, D.}, \bibinfo{author}{Zhou, L.}, \bibinfo{year}{2023}.
\newblock \bibinfo{title}{Q2atransformer: Improving medical vqa via an answer querying decoder}, in: \bibinfo{booktitle}{International Conference on Information Processing in Medical Imaging}, \bibinfo{organization}{Springer}. pp. \bibinfo{pages}{445--456}.
%Type = Inproceedings
\bibitem[{Lu et~al.(2017)Lu, Xiong, Parikh and Socher}]{2017Knowing}
\bibinfo{author}{Lu, J.}, \bibinfo{author}{Xiong, C.}, \bibinfo{author}{Parikh, D.}, \bibinfo{author}{Socher, R.}, \bibinfo{year}{2017}.
\newblock \bibinfo{title}{Knowing when to look: Adaptive attention via a visual sentinel for image captioning}, in: \bibinfo{booktitle}{2017 {IEEE} Conference on Computer Vision and Pattern Recognition, {CVPR} 2017, Honolulu, HI, USA, July 21-26, 2017}, \bibinfo{publisher}{{IEEE} Computer Society}. pp. \bibinfo{pages}{3242--3250}.
\newblock \URLprefix \url{https://doi.org/10.1109/CVPR.2017.345}, \DOIprefix\doi{10.1109/CVPR.2017.345}.
%Type = Inproceedings
\bibitem[{Nguyen et~al.(2019)Nguyen, Do, Nguyen, Do, Tjiputra and Tran}]{nguyen2019overcoming}
\bibinfo{author}{Nguyen, B.D.}, \bibinfo{author}{Do, T.T.}, \bibinfo{author}{Nguyen, B.X.}, \bibinfo{author}{Do, T.}, \bibinfo{author}{Tjiputra, E.}, \bibinfo{author}{Tran, Q.D.}, \bibinfo{year}{2019}.
\newblock \bibinfo{title}{Overcoming data limitation in medical visual question answering}, in: \bibinfo{booktitle}{International Conference on Medical Image Computing and Computer-Assisted Intervention}, \bibinfo{organization}{Springer}. pp. \bibinfo{pages}{522--530}.
%Type = Article
\bibitem[{OpenAI(2023)}]{OpenAI2023GPT4TR}
\bibinfo{author}{OpenAI}, \bibinfo{year}{2023}.
\newblock \bibinfo{title}{Gpt-4 technical report}.
\newblock \bibinfo{journal}{ArXiv} \bibinfo{volume}{abs/2303.08774}.
\newblock \URLprefix \url{https://api.semanticscholar.org/CorpusID:257532815}.
%Type = Inproceedings
\bibitem[{Pan et~al.(2020)Pan, Yao, Li and Mei}]{pan2020xlinear}
\bibinfo{author}{Pan, Y.}, \bibinfo{author}{Yao, T.}, \bibinfo{author}{Li, Y.}, \bibinfo{author}{Mei, T.}, \bibinfo{year}{2020}.
\newblock \bibinfo{title}{X-linear attention networks for image captioning}, in: \bibinfo{booktitle}{2020 {IEEE/CVF} Conference on Computer Vision and Pattern Recognition, {CVPR} 2020, Seattle, WA, USA, June 13-19, 2020}, \bibinfo{publisher}{Computer Vision Foundation / {IEEE}}. pp. \bibinfo{pages}{10968--10977}.
\newblock \URLprefix \url{https://openaccess.thecvf.com/content\_CVPR\_2020/html/Pan\_X-Linear\_Attention\_Networks\_for\_Image\_Captioning\_CVPR\_2020\_paper.html}, \DOIprefix\doi{10.1109/CVPR42600.2020.01098}.
%Type = Inproceedings
\bibitem[{Papineni et~al.(2002)Papineni, Roukos, Ward and Zhu}]{Kishore2002bleu}
\bibinfo{author}{Papineni, K.}, \bibinfo{author}{Roukos, S.}, \bibinfo{author}{Ward, T.}, \bibinfo{author}{Zhu, W.}, \bibinfo{year}{2002}.
\newblock \bibinfo{title}{Bleu: a method for automatic evaluation of machine translation}, in: \bibinfo{booktitle}{Proceedings of the 40th Annual Meeting of the Association for Computational Linguistics, July 6-12, 2002, Philadelphia, PA, {USA}}, \bibinfo{publisher}{{ACL}}. pp. \bibinfo{pages}{311--318}.
\newblock \URLprefix \url{https://aclanthology.org/P02-1040/}, \DOIprefix\doi{10.3115/1073083.1073135}.
%Type = Inproceedings
\bibitem[{Pellegrini et~al.(2023)Pellegrini, Keicher, {\"O}zsoy and Navab}]{pellegrini2023rad}
\bibinfo{author}{Pellegrini, C.}, \bibinfo{author}{Keicher, M.}, \bibinfo{author}{{\"O}zsoy, E.}, \bibinfo{author}{Navab, N.}, \bibinfo{year}{2023}.
\newblock \bibinfo{title}{Rad-restruct: A novel vqa benchmark and method for structured radiology reporting}, in: \bibinfo{booktitle}{International Conference on Medical Image Computing and Computer-Assisted Intervention}, \bibinfo{organization}{Springer}. pp. \bibinfo{pages}{409--419}.
%Type = Article
\bibitem[{Peng et~al.(2023)Peng, Wang, Dong, Hao, Huang, Ma and Wei}]{peng2023kosmos}
\bibinfo{author}{Peng, Z.}, \bibinfo{author}{Wang, W.}, \bibinfo{author}{Dong, L.}, \bibinfo{author}{Hao, Y.}, \bibinfo{author}{Huang, S.}, \bibinfo{author}{Ma, S.}, \bibinfo{author}{Wei, F.}, \bibinfo{year}{2023}.
\newblock \bibinfo{title}{Kosmos-2: Grounding multimodal large language models to the world}.
\newblock \bibinfo{journal}{arXiv preprint arXiv:2306.14824} .
%Type = Article
\bibitem[{Shi et~al.(2023)Shi, Peng, Liao, Lin, Chen, Liu, Zhang and Jin}]{shi2023exploring}
\bibinfo{author}{Shi, Y.}, \bibinfo{author}{Peng, D.}, \bibinfo{author}{Liao, W.}, \bibinfo{author}{Lin, Z.}, \bibinfo{author}{Chen, X.}, \bibinfo{author}{Liu, C.}, \bibinfo{author}{Zhang, Y.}, \bibinfo{author}{Jin, L.}, \bibinfo{year}{2023}.
\newblock \bibinfo{title}{Exploring ocr capabilities of gpt-4v (ision): A quantitative and in-depth evaluation}.
\newblock \bibinfo{journal}{arXiv preprint arXiv:2310.16809} .
%Type = Article
\bibitem[{Singhal et~al.(2023)Singhal, Tu, Gottweis, Sayres, Wulczyn, Hou, Clark, Pfohl, Cole-Lewis, Neal et~al.}]{singhal2023towards}
\bibinfo{author}{Singhal, K.}, \bibinfo{author}{Tu, T.}, \bibinfo{author}{Gottweis, J.}, \bibinfo{author}{Sayres, R.}, \bibinfo{author}{Wulczyn, E.}, \bibinfo{author}{Hou, L.}, \bibinfo{author}{Clark, K.}, \bibinfo{author}{Pfohl, S.}, \bibinfo{author}{Cole-Lewis, H.}, \bibinfo{author}{Neal, D.}, et~al., \bibinfo{year}{2023}.
\newblock \bibinfo{title}{Towards expert-level medical question answering with large language models}.
\newblock \bibinfo{journal}{arXiv preprint arXiv:2305.09617} .
%Type = Inproceedings
\bibitem[{Sun et~al.(2023a)Sun, Wei, Xu, Lu, Liu, Wang and Zheng}]{sun2023you}
\bibinfo{author}{Sun, J.}, \bibinfo{author}{Wei, D.}, \bibinfo{author}{Xu, Z.}, \bibinfo{author}{Lu, D.}, \bibinfo{author}{Liu, H.}, \bibinfo{author}{Wang, L.}, \bibinfo{author}{Zheng, Y.}, \bibinfo{year}{2023}a.
\newblock \bibinfo{title}{You’ve got two teachers: Co-evolutionary image and report distillation for semi-supervised anatomical abnormality detection in chest x-ray}, in: \bibinfo{booktitle}{International Conference on Medical Image Computing and Computer-Assisted Intervention}, \bibinfo{organization}{Springer}. pp. \bibinfo{pages}{363--373}.
%Type = Article
\bibitem[{Sun et~al.(2023b)Sun, Lin, Zhu, Xie, Wang, Lu and Peng}]{sun2023scoping}
\bibinfo{author}{Sun, Z.}, \bibinfo{author}{Lin, M.}, \bibinfo{author}{Zhu, Q.}, \bibinfo{author}{Xie, Q.}, \bibinfo{author}{Wang, F.}, \bibinfo{author}{Lu, Z.}, \bibinfo{author}{Peng, Y.}, \bibinfo{year}{2023}b.
\newblock \bibinfo{title}{A scoping review on multimodal deep learning in biomedical images and texts}.
\newblock \bibinfo{journal}{Journal of Biomedical Informatics} , \bibinfo{pages}{104482}.
%Type = Article
\bibitem[{Touvron et~al.(2023)Touvron, Lavril, Izacard, Martinet, Lachaux, Lacroix, Rozi{\`e}re, Goyal, Hambro, Azhar et~al.}]{touvron2023llama}
\bibinfo{author}{Touvron, H.}, \bibinfo{author}{Lavril, T.}, \bibinfo{author}{Izacard, G.}, \bibinfo{author}{Martinet, X.}, \bibinfo{author}{Lachaux, M.A.}, \bibinfo{author}{Lacroix, T.}, \bibinfo{author}{Rozi{\`e}re, B.}, \bibinfo{author}{Goyal, N.}, \bibinfo{author}{Hambro, E.}, \bibinfo{author}{Azhar, F.}, et~al., \bibinfo{year}{2023}.
\newblock \bibinfo{title}{Llama: Open and efficient foundation language models}.
\newblock \bibinfo{journal}{arXiv preprint arXiv:2302.13971} .
%Type = Article
\bibitem[{Tsimpoukelli et~al.(2021)Tsimpoukelli, Menick, Cabi, Eslami, Vinyals and Hill}]{tsimpoukelli2021multimodal}
\bibinfo{author}{Tsimpoukelli, M.}, \bibinfo{author}{Menick, J.L.}, \bibinfo{author}{Cabi, S.}, \bibinfo{author}{Eslami, S.}, \bibinfo{author}{Vinyals, O.}, \bibinfo{author}{Hill, F.}, \bibinfo{year}{2021}.
\newblock \bibinfo{title}{Multimodal few-shot learning with frozen language models}.
\newblock \bibinfo{journal}{Advances in Neural Information Processing Systems} \bibinfo{volume}{34}, \bibinfo{pages}{200--212}.
%Type = Inproceedings
\bibitem[{Vaswani et~al.(2017)Vaswani, Shazeer, Parmar, Uszkoreit, Jones, Gomez, Kaiser and Polosukhin}]{vaswani2017attentionisallyouneed}
\bibinfo{author}{Vaswani, A.}, \bibinfo{author}{Shazeer, N.}, \bibinfo{author}{Parmar, N.}, \bibinfo{author}{Uszkoreit, J.}, \bibinfo{author}{Jones, L.}, \bibinfo{author}{Gomez, A.N.}, \bibinfo{author}{Kaiser, L.}, \bibinfo{author}{Polosukhin, I.}, \bibinfo{year}{2017}.
\newblock \bibinfo{title}{Attention is all you need}, in: \bibinfo{editor}{Guyon, I.}, \bibinfo{editor}{von Luxburg, U.}, \bibinfo{editor}{Bengio, S.}, \bibinfo{editor}{Wallach, H.M.}, \bibinfo{editor}{Fergus, R.}, \bibinfo{editor}{Vishwanathan, S.V.N.}, \bibinfo{editor}{Garnett, R.} (Eds.), \bibinfo{booktitle}{Advances in Neural Information Processing Systems 30: Annual Conference on Neural Information Processing Systems 2017, December 4-9, 2017, Long Beach, CA, {USA}}, pp. \bibinfo{pages}{5998--6008}.
\newblock \URLprefix \url{https://proceedings.neurips.cc/paper/2017/hash/3f5ee243547dee91fbd053c1c4a845aa-Abstract.html}.
%Type = Inproceedings
\bibitem[{Vedantam et~al.(2015)Vedantam, Zitnick and Parikh}]{vedantam2015cider}
\bibinfo{author}{Vedantam, R.}, \bibinfo{author}{Zitnick, C.L.}, \bibinfo{author}{Parikh, D.}, \bibinfo{year}{2015}.
\newblock \bibinfo{title}{Cider: Consensus-based image description evaluation}, in: \bibinfo{booktitle}{{IEEE} Conference on Computer Vision and Pattern Recognition, {CVPR} 2015, Boston, MA, USA, June 7-12, 2015}, \bibinfo{publisher}{{IEEE} Computer Society}. pp. \bibinfo{pages}{4566--4575}.
\newblock \URLprefix \url{https://doi.org/10.1109/CVPR.2015.7299087}, \DOIprefix\doi{10.1109/CVPR.2015.7299087}.
%Type = Inproceedings
\bibitem[{Vinyals et~al.(2015)Vinyals, Toshev, Bengio and Erhan}]{vinyals2015showandtell}
\bibinfo{author}{Vinyals, O.}, \bibinfo{author}{Toshev, A.}, \bibinfo{author}{Bengio, S.}, \bibinfo{author}{Erhan, D.}, \bibinfo{year}{2015}.
\newblock \bibinfo{title}{Show and tell: {A} neural image caption generator}, in: \bibinfo{booktitle}{{IEEE} Conference on Computer Vision and Pattern Recognition, {CVPR} 2015, Boston, MA, USA, June 7-12, 2015}, \bibinfo{publisher}{{IEEE} Computer Society}. pp. \bibinfo{pages}{3156--3164}.
\newblock \URLprefix \url{https://doi.org/10.1109/CVPR.2015.7298935}, \DOIprefix\doi{10.1109/CVPR.2015.7298935}.
%Type = Article
\bibitem[{Wang et~al.(2022a)Wang, Wei, Schuurmans, Le, Chi, Narang, Chowdhery and Zhou}]{wang2022self}
\bibinfo{author}{Wang, X.}, \bibinfo{author}{Wei, J.}, \bibinfo{author}{Schuurmans, D.}, \bibinfo{author}{Le, Q.}, \bibinfo{author}{Chi, E.}, \bibinfo{author}{Narang, S.}, \bibinfo{author}{Chowdhery, A.}, \bibinfo{author}{Zhou, D.}, \bibinfo{year}{2022}a.
\newblock \bibinfo{title}{Self-consistency improves chain of thought reasoning in language models}.
\newblock \bibinfo{journal}{arXiv preprint arXiv:2203.11171} .
%Type = Inproceedings
\bibitem[{Wang et~al.(2023a)Wang, Liu, Wang and Zhou}]{wang2023metransformer}
\bibinfo{author}{Wang, Z.}, \bibinfo{author}{Liu, L.}, \bibinfo{author}{Wang, L.}, \bibinfo{author}{Zhou, L.}, \bibinfo{year}{2023}a.
\newblock \bibinfo{title}{Metransformer: Radiology report generation by transformer with multiple learnable expert tokens}, in: \bibinfo{booktitle}{{IEEE/CVF} Conference on Computer Vision and Pattern Recognition, {CVPR} 2023, Vancouver, BC, Canada, June 17-24, 2023}, \bibinfo{publisher}{{IEEE}}. pp. \bibinfo{pages}{11558--11567}.
\newblock \URLprefix \url{https://doi.org/10.1109/CVPR52729.2023.01112}, \DOIprefix\doi{10.1109/CVPR52729.2023.01112}.
%Type = Article
\bibitem[{Wang et~al.(2023b)Wang, Liu, Wang and Zhou}]{wang2023r2gengpt}
\bibinfo{author}{Wang, Z.}, \bibinfo{author}{Liu, L.}, \bibinfo{author}{Wang, L.}, \bibinfo{author}{Zhou, L.}, \bibinfo{year}{2023}b.
\newblock \bibinfo{title}{R2gengpt: Radiology report generation with frozen llms}.
\newblock \bibinfo{journal}{arXiv preprint arXiv:2309.09812} .
%Type = Inproceedings
\bibitem[{Wang et~al.(2022b)Wang, Tang, Wang, Li and Zhou}]{wang2022medical}
\bibinfo{author}{Wang, Z.}, \bibinfo{author}{Tang, M.}, \bibinfo{author}{Wang, L.}, \bibinfo{author}{Li, X.}, \bibinfo{author}{Zhou, L.}, \bibinfo{year}{2022}b.
\newblock \bibinfo{title}{A medical semantic-assisted transformer for radiographic report generation}, in: \bibinfo{editor}{Wang, L.}, \bibinfo{editor}{Dou, Q.}, \bibinfo{editor}{Fletcher, P.T.}, \bibinfo{editor}{Speidel, S.}, \bibinfo{editor}{Li, S.} (Eds.), \bibinfo{booktitle}{Medical Image Computing and Computer Assisted Intervention - {MICCAI} 2022 - 25th International Conference, Singapore, September 18-22, 2022, Proceedings, Part {III}}, \bibinfo{publisher}{Springer}. pp. \bibinfo{pages}{655--664}.
\newblock \URLprefix \url{https://doi.org/10.1007/978-3-031-16437-8\_63}, \DOIprefix\doi{10.1007/978-3-031-16437-8\_63}.
%Type = Article
\bibitem[{Wei et~al.(2022a)Wei, Tay, Bommasani, Raffel, Zoph, Borgeaud, Yogatama, Bosma, Zhou, Metzler et~al.}]{wei2022emergent}
\bibinfo{author}{Wei, J.}, \bibinfo{author}{Tay, Y.}, \bibinfo{author}{Bommasani, R.}, \bibinfo{author}{Raffel, C.}, \bibinfo{author}{Zoph, B.}, \bibinfo{author}{Borgeaud, S.}, \bibinfo{author}{Yogatama, D.}, \bibinfo{author}{Bosma, M.}, \bibinfo{author}{Zhou, D.}, \bibinfo{author}{Metzler, D.}, et~al., \bibinfo{year}{2022}a.
\newblock \bibinfo{title}{Emergent abilities of large language models}.
\newblock \bibinfo{journal}{arXiv preprint arXiv:2206.07682} .
%Type = Article
\bibitem[{Wei et~al.(2022b)Wei, Wang, Schuurmans, Bosma, Xia, Chi, Le, Zhou et~al.}]{wei2022chain}
\bibinfo{author}{Wei, J.}, \bibinfo{author}{Wang, X.}, \bibinfo{author}{Schuurmans, D.}, \bibinfo{author}{Bosma, M.}, \bibinfo{author}{Xia, F.}, \bibinfo{author}{Chi, E.}, \bibinfo{author}{Le, Q.V.}, \bibinfo{author}{Zhou, D.}, et~al., \bibinfo{year}{2022}b.
\newblock \bibinfo{title}{Chain-of-thought prompting elicits reasoning in large language models}.
\newblock \bibinfo{journal}{Advances in Neural Information Processing Systems} \bibinfo{volume}{35}, \bibinfo{pages}{24824--24837}.
%Type = Article
\bibitem[{Wu et~al.(2023a)Wu, Lei, Zheng, Zhao, Lin, Zhang, Zhou, Zhao, Zhang, Wang et~al.}]{wu2023can}
\bibinfo{author}{Wu, C.}, \bibinfo{author}{Lei, J.}, \bibinfo{author}{Zheng, Q.}, \bibinfo{author}{Zhao, W.}, \bibinfo{author}{Lin, W.}, \bibinfo{author}{Zhang, X.}, \bibinfo{author}{Zhou, X.}, \bibinfo{author}{Zhao, Z.}, \bibinfo{author}{Zhang, Y.}, \bibinfo{author}{Wang, Y.}, et~al., \bibinfo{year}{2023}a.
\newblock \bibinfo{title}{Can gpt-4v (ision) serve medical applications? case studies on gpt-4v for multimodal medical diagnosis}.
\newblock \bibinfo{journal}{arXiv preprint arXiv:2310.09909} .
%Type = Inproceedings
\bibitem[{Wu et~al.(2019)Wu, Liu, Wang and Li}]{wu2019differential}
\bibinfo{author}{Wu, C.}, \bibinfo{author}{Liu, J.}, \bibinfo{author}{Wang, X.}, \bibinfo{author}{Li, R.}, \bibinfo{year}{2019}.
\newblock \bibinfo{title}{Differential networks for visual question answering}, in: \bibinfo{booktitle}{Proceedings of the AAAI Conference on Artificial Intelligence}, pp. \bibinfo{pages}{8997--9004}.
%Type = Article
\bibitem[{Wu et~al.(2023b)Wu, Wang, Yang, Zheng, Zhang, Zhao and Qin}]{wu2023early}
\bibinfo{author}{Wu, Y.}, \bibinfo{author}{Wang, S.}, \bibinfo{author}{Yang, H.}, \bibinfo{author}{Zheng, T.}, \bibinfo{author}{Zhang, H.}, \bibinfo{author}{Zhao, Y.}, \bibinfo{author}{Qin, B.}, \bibinfo{year}{2023}b.
\newblock \bibinfo{title}{An early evaluation of gpt-4v (ision)}.
\newblock \bibinfo{journal}{arXiv preprint arXiv:2310.16534} .
%Type = Inproceedings
\bibitem[{Xu et~al.(2015)Xu, Ba, Kiros, Cho, Courville, Salakhutdinov, Zemel and Bengio}]{xu2016showattandtell}
\bibinfo{author}{Xu, K.}, \bibinfo{author}{Ba, J.}, \bibinfo{author}{Kiros, R.}, \bibinfo{author}{Cho, K.}, \bibinfo{author}{Courville, A.C.}, \bibinfo{author}{Salakhutdinov, R.}, \bibinfo{author}{Zemel, R.S.}, \bibinfo{author}{Bengio, Y.}, \bibinfo{year}{2015}.
\newblock \bibinfo{title}{Show, attend and tell: Neural image caption generation with visual attention}, in: \bibinfo{editor}{Bach, F.R.}, \bibinfo{editor}{Blei, D.M.} (Eds.), \bibinfo{booktitle}{Proceedings of the 32nd International Conference on Machine Learning, {ICML} 2015, Lille, France, 6-11 July 2015}, \bibinfo{publisher}{JMLR.org}. pp. \bibinfo{pages}{2048--2057}.
\newblock \URLprefix \url{http://proceedings.mlr.press/v37/xuc15.html}.
%Type = Article
\bibitem[{Yang et~al.(2023a)Yang, Zhang, Li, Zou, Li and Gao}]{yang2023set}
\bibinfo{author}{Yang, J.}, \bibinfo{author}{Zhang, H.}, \bibinfo{author}{Li, F.}, \bibinfo{author}{Zou, X.}, \bibinfo{author}{Li, C.}, \bibinfo{author}{Gao, J.}, \bibinfo{year}{2023}a.
\newblock \bibinfo{title}{Set-of-mark prompting unleashes extraordinary visual grounding in gpt-4v}.
\newblock \bibinfo{journal}{arXiv preprint arXiv:2310.11441} .
%Type = Article
\bibitem[{Yang et~al.(2021)Yang, Wu, Ge, Zhou and Xiao}]{2021Knowledge}
\bibinfo{author}{Yang, S.}, \bibinfo{author}{Wu, X.}, \bibinfo{author}{Ge, S.}, \bibinfo{author}{Zhou, S.K.}, \bibinfo{author}{Xiao, L.}, \bibinfo{year}{2021}.
\newblock \bibinfo{title}{Knowledge matters: Radiology report generation with general and specific knowledge}.
\newblock \bibinfo{journal}{Medical Image Analysis} .
%Type = Article
\bibitem[{Yang et~al.(2023b)Yang, Li, Lin, Wang, Lin, Liu and Wang}]{yang2023dawn}
\bibinfo{author}{Yang, Z.}, \bibinfo{author}{Li, L.}, \bibinfo{author}{Lin, K.}, \bibinfo{author}{Wang, J.}, \bibinfo{author}{Lin, C.C.}, \bibinfo{author}{Liu, Z.}, \bibinfo{author}{Wang, L.}, \bibinfo{year}{2023}b.
\newblock \bibinfo{title}{The dawn of lmms: Preliminary explorations with gpt-4v (ision)}.
\newblock \bibinfo{journal}{arXiv preprint arXiv:2309.17421} .
%Type = Article
\bibitem[{Ye et~al.(2023)Ye, Xu, Xu, Ye, Yan, Zhou, Wang, Hu, Shi, Shi et~al.}]{ye2023mplug}
\bibinfo{author}{Ye, Q.}, \bibinfo{author}{Xu, H.}, \bibinfo{author}{Xu, G.}, \bibinfo{author}{Ye, J.}, \bibinfo{author}{Yan, M.}, \bibinfo{author}{Zhou, Y.}, \bibinfo{author}{Wang, J.}, \bibinfo{author}{Hu, A.}, \bibinfo{author}{Shi, P.}, \bibinfo{author}{Shi, Y.}, et~al., \bibinfo{year}{2023}.
\newblock \bibinfo{title}{mplug-owl: Modularization empowers large language models with multimodality}.
\newblock \bibinfo{journal}{arXiv preprint arXiv:2304.14178} .
%Type = Article
\bibitem[{Zhang et~al.(2020)Zhang, Wang, Xu, Yu, Yuille and Xu}]{2020When}
\bibinfo{author}{Zhang, Y.}, \bibinfo{author}{Wang, X.}, \bibinfo{author}{Xu, Z.}, \bibinfo{author}{Yu, Q.}, \bibinfo{author}{Yuille, A.}, \bibinfo{author}{Xu, D.}, \bibinfo{year}{2020}.
\newblock \bibinfo{title}{When radiology report generation meets knowledge graph}.
\newblock \bibinfo{journal}{Proceedings of the AAAI Conference on Artificial Intelligence} .
%Type = Article
\bibitem[{Zhao et~al.(2023)Zhao, Lin, Zhou, Huang, Feng and Kang}]{zhao2023bubogpt}
\bibinfo{author}{Zhao, Y.}, \bibinfo{author}{Lin, Z.}, \bibinfo{author}{Zhou, D.}, \bibinfo{author}{Huang, Z.}, \bibinfo{author}{Feng, J.}, \bibinfo{author}{Kang, B.}, \bibinfo{year}{2023}.
\newblock \bibinfo{title}{Bubogpt: Enabling visual grounding in multi-modal llms}.
\newblock \bibinfo{journal}{arXiv preprint arXiv:2307.08581} .
%Type = Inproceedings
\bibitem[{Zhu et~al.(2022)Zhu, Zhou, Shen, Luo, Pan, Lin, Chen, Cao, Sun and Ji}]{zhu2022seqtr}
\bibinfo{author}{Zhu, C.}, \bibinfo{author}{Zhou, Y.}, \bibinfo{author}{Shen, Y.}, \bibinfo{author}{Luo, G.}, \bibinfo{author}{Pan, X.}, \bibinfo{author}{Lin, M.}, \bibinfo{author}{Chen, C.}, \bibinfo{author}{Cao, L.}, \bibinfo{author}{Sun, X.}, \bibinfo{author}{Ji, R.}, \bibinfo{year}{2022}.
\newblock \bibinfo{title}{Seqtr: A simple yet universal network for visual grounding}, in: \bibinfo{booktitle}{European Conference on Computer Vision}, \bibinfo{organization}{Springer}. pp. \bibinfo{pages}{598--615}.

\end{thebibliography}

\newpage
\appendix
\section{Appendix}

\subsection{Details of Prompt Settings}
\label{sec:Prompt Settings}
To elucidate our prompt design and selection process, we present illustrative cases for better comprehension. We showcase instances involving both zero-shot and few-shot prompts to elucidate the distinction.  In the zero-shot scenario, the model relies solely on its inherent training and knowledge base. Conversely, the few-shot scenario incorporates specific examples, enhancing the model's understanding and response accuracy. 
Our few-shot prompt encompasses diverse cases utilising both normal and abnormal examples, showcasing radiology images with normal and abnormal findings. %In the few-shot prompt case, we give different cases by using normal examples and abnormal examples, which means the report shows the radiology image is normal and abnormal. 
In all prompts, we prompt GPT-4V to assume the role of a professional radiologist. Additionally, we explicitly instruct it to generate both the impression and findings sections.

\subsubsection{Zero-shot prompt} 

\begin{figure}[h]
\begin{center}
    \includegraphics[width=0.5\textwidth]{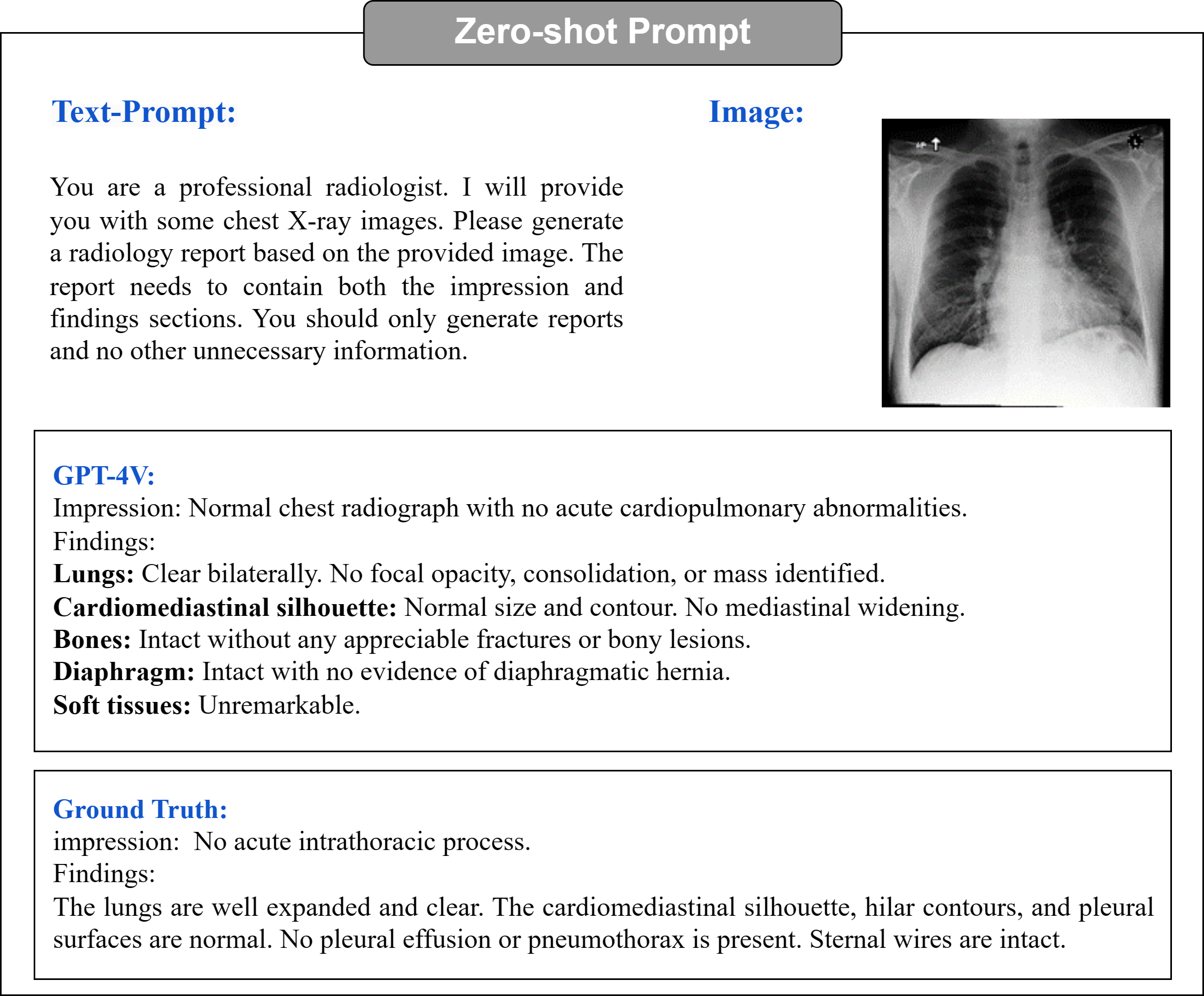}
\end{center}
\caption{Zero-shot prompt. No additional information provided to GPT-4V.}
\label{fig:Zero shot prompt example}
\end{figure}

Figure~\ref{fig:Zero shot prompt example} showcases a zero-shot prompt example. We did not add any additional information to the text prompt.

% \begin{figure}[h]
% \includegraphics[width=\textwidth]{R2Gen_Prompts/Zero shot prompt.png}
% \caption{Zero shot prompt}
% \label{fig:Zero shot prompt}
% \end{figure}

\subsubsection{Few-shot prompt} 
\label{sec:Few-shot prompt}
\begin{figure}[]
\begin{center}
%\framebox[4.0in]{$\;$}
\includegraphics[width=0.5\textwidth]{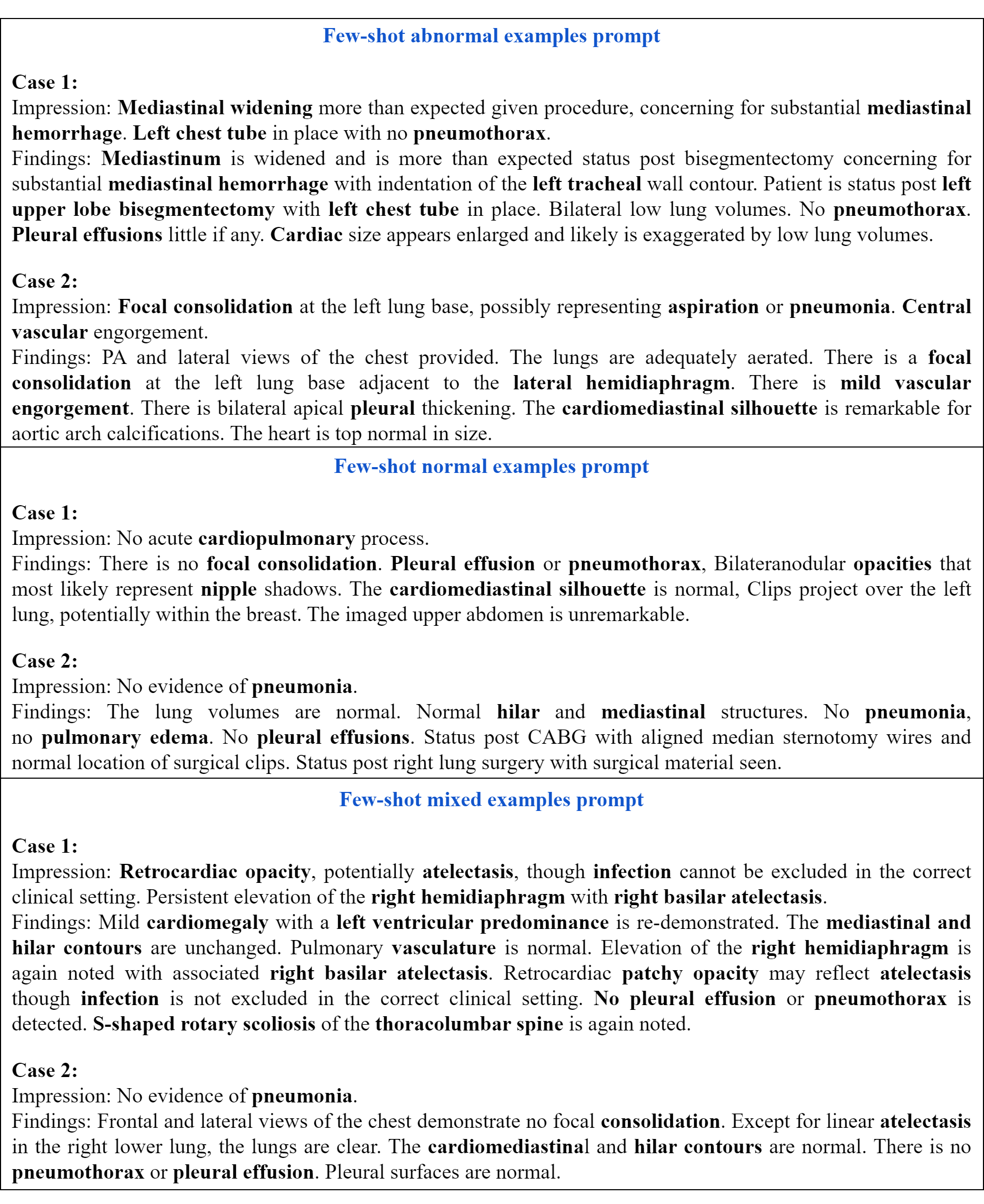}
\end{center}
\caption{Example reports in prompts: Three pairs of different example reports in few-shot prompt settings. We added these example reports to few-shot prompts to help GPT-4V generate radiology reports.}
\label{fig:Example reports}
\end{figure}

We add two example reports to the prompt while exploring three different combinations: (1) exclusively using normal examples, (2) exclusively using abnormal examples), (3) combining one normal and one abnormal example. The example reports are given in Figure~\ref{fig:Example reports}.
% \begin{figure}[h]
% \begin{center}
%     \includegraphics[width=0.5\textwidth]{R2Gen_Prompts/few_shots_prompt.png}
% \end{center}
% \caption{Few-shot prompt. Example reports from MIMIC-CXR training dataset are added to the prompt text}
% \label{fig:Few shots prompt}
% \end{figure}

\noindent \textbf{Few-shot normal examples prompt} In this prompt method, we curated reports from two normal samples within the MIMIC-CXR training set. To ensure comprehensiveness, we specifically chose reports with rich content.

\noindent \textbf{Few-shot abnormal examples prompt} In this prompt method, we carefully chose two reports originating from abnormal samples within the MIMIC-CXR training set.

\noindent \textbf{Few-shot mixed-example prompt} In this prompt method, we chose one normal and one abnormal report from the MIMIC-CXR training set. The sequence in which these two examples are presented is not anticipated to significantly impact the generated results. In this specific experiment, we positioned the abnormal report before the normal one. 

% \subsection{Incorrect viewpoint case}
% \label{sec:Incorrect viewpoint case}
% In Figure~\ref{fig:Viewpoint information Case 2}, it becomes evident that the chest X-ray image provided is a frontal view, whereas GPT-4V's generated report incorrectly labels it as a lateral view.

% \begin{figure}[h]
% \begin{center}
% %\framebox[4.0in]{$\;$}
% \includegraphics[width=0.5\textwidth]{R2Gen_case/Viewpoints_2.png}
% \end{center}
% \caption{Viewpoint information Case 2. While GPT-4V provides view information, it is inaccurate.}
% \label{fig:Viewpoint information Case 2}
% \end{figure}

\end{document}